\documentclass[10pt,twocolumn,letterpaper]{article}

\usepackage[pagenumbers]{cvpr} %

\usepackage{cuted} %

\usepackage{currfile} %

\usepackage{caption} %

\AtBeginDocument{\setlength{\parskip}{0pt}}
\newcommand{\rhead}[1]{\par\addvspace{3pt}\noindent\textbf{#1}\enspace\ignorespaces}

\usepackage{amsmath,amsthm,amssymb}
\usepackage{booktabs,multirow,array}
\usepackage{graphicx}
\usepackage{tikz}
\usetikzlibrary{positioning,arrows.meta,fit,backgrounds,calc}
\usepackage{xcolor}
\usepackage{colortbl}
\definecolor{goodgreen}{HTML}{E4F3EA}
\definecolor{badred}{HTML}{FBE7EA}
\newcommand{\good}{\cellcolor{goodgreen}}
\newcommand{\bad}{\cellcolor{badred}}
\newcommand{\ours}{\rowcolor{black!8}}
\newcommand{\zerogood}[1]{\ifnum#1=0\relax\cellcolor{goodgreen}\else\cellcolor{badred}\fi#1}
\usepackage{ifthen}
\usepackage{xspace}
\usepackage{enumitem}
\definecolor{pendingred}{HTML}{c0392b}

\definecolor{cvprblue}{rgb}{0.21,0.49,0.74}
\usepackage[pagebackref,breaklinks,colorlinks,allcolors=cvprblue]{hyperref}
\usepackage{algorithm}
\usepackage{algpseudocode}
\usepackage[capitalize,noabbrev]{cleveref}
\AtBeginDocument{\crefname{figure}{Fig.}{Figs.}}

\newtheorem{proposition}{Proposition}
\newcommand{\method}{{\normalfont\textsc{SealedLoop}}\xspace}
\newcommand{\Jin}{J_{\mathrm{in}}}
\newcommand{\Jout}{J_{\mathrm{out}}}
\newcommand{\Din}{\mathcal{D}_{\mathrm{in}}}
\newcommand{\Pool}{\mathcal{P}}
\newcommand{\supp}[1]{Appendix~\ref{\csname suppsec@#1\endcsname}}
\makeatletter
\@namedef{suppsec@Method details}{sec:supp_method}\@namedef{suppsec@Manipulation}{sec:supp_manip}\@namedef{suppsec@Driving}{sec:supp_drive}
\@namedef{suppsec@Real-world field experiment}{sec:supp_vehicle}\@namedef{suppsec@Survey}{sec:supp_survey}\@namedef{suppsec@Additional analyses}{sec:supp_additional}
\makeatother

\IfFileExists{sl_result_macros.tex}{
\newcommand{\SLSealedAcceptedGensSThree}{0}
\newcommand{\SLSealedRollbacksSThree}{15}
\newcommand{\SLSealedGainFirstMeanSThree}{\ensuremath{-0.029}}
\newcommand{\SLSealedChainsSThree}{3}
\newcommand{\SLInLoopAcceptedGensSThree}{3}
\newcommand{\SLInLoopAcceptedWorseSThree}{3}
\newcommand{\SLInLoopChainsSThree}{6}
\newcommand{\SLFalseRollbacksSThree}{0}
\newcommand{\SLSealedAcceptedGensSOne}{12}
\newcommand{\SLSealedGainLastSOne}{\ensuremath{+0.147}}
\newcommand{\SLSealedChainsSOne}{9}
\newcommand{\SLDivergenceGenNumSOne}{3}
\newcommand{\SLInLoopAcceptedGensSOne}{13}
\newcommand{\SLInLoopAcceptedWorseSOne}{1}
\newcommand{\SLInLoopChainsSOne}{9}
\newcommand{\SLFalseRollbacksSOne}{3}
\providecommand{\SLEvalPerGen}{\SLPending}
\providecommand{\SLTasksSOne}{\SLPending}
\providecommand{\SLSealedAcceptedGensSOne}{\SLPending}
\providecommand{\SLSealedGainLastSOne}{\SLPending}
\providecommand{\SLSealedChainsSOne}{\SLPending}
\providecommand{\SLDivergenceGenNumSOne}{\SLPending}
\providecommand{\SLInLoopAcceptedGensSOne}{\SLPending}
\providecommand{\SLInLoopAcceptedWorseSOne}{\SLPending}
\providecommand{\SLInLoopChainsSOne}{\SLPending}
\providecommand{\SLFalseRollbacksSOne}{\SLPending}
\providecommand{\SLSealedAcceptedGensSThree}{\SLPending}
\providecommand{\SLSealedRollbacksSThree}{\SLPending}
\providecommand{\SLSealedChainsSThree}{\SLPending}
\providecommand{\SLInLoopAcceptedGensSThree}{\SLPending}
\providecommand{\SLInLoopAcceptedWorseSThree}{\SLPending}
\providecommand{\SLInLoopChainsSThree}{\SLPending}
\providecommand{\SLFalseRollbacksSThree}{\SLPending}
\providecommand{\SLStartsInBand}{\SLPending}
\providecommand{\SLExtendedChainsSOne}{\SLPending}
\providecommand{\SLDistinctTrajectoriesSOne}{\SLPending}
\providecommand{\SLSeedOutcomeAgreementSOne}{\SLPending}
\providecommand{\SLPairedOutcomeIdenticalSOne}{\SLPending}
}{\input{sl_pending_macros.tex}}
\let\SLTasksSOne\undefined
\let\SLExtendedChainsSOne\undefined

\let\SLDistinctTrajectoriesSOne\undefined
\let\SLSeedOutcomeAgreementSOne\undefined
\let\SLPairedOutcomeIdenticalSOne\undefined
\newcommand{\SLCriticSelTrainSThree}{\ensuremath{+0.010}} %
\newcommand{\SLCriticSelHeldSThree}{\ensuremath{-0.042}} %
\newcommand{\SLCriticFPRTrainSThree}{0.164} %
\newcommand{\SLCriticFPRHeldSThree}{0.383} %
\newcommand{\SLCriticSelTrainCtlSThree}{\ensuremath{+0.001}} %
\newcommand{\SLCriticSelHeldCtlSThree}{\ensuremath{-0.073}} %
\newcommand{\SLJudgeGapGenOneInLoopROneSThree}{\ensuremath{+0.051}} %
\newcommand{\SLJudgeGapGenOneCIInLoopROneSThree}{[\ensuremath{+0.045}, \ensuremath{+0.057}]} %
\newcommand{\SLJudgeGapGenOneInLoopRTwoSThree}{\ensuremath{+0.091}} %
\newcommand{\SLJudgeGapGenOneCIInLoopRTwoSThree}{[\ensuremath{+0.084}, \ensuremath{+0.098}]} %
\newcommand{\SLFTwoHitRateSThree}{17/45} %
\newcommand{\SLFTwoPriorBandThroughoutSThree}{19/45} %
\newcommand{\SLBlockMDEPTwoSThree}{0.030} %
\newcommand{\SLVinReplayFrameAcceptCtlSThree}{15/15} %
\newcommand{\SLVinReplayBlockAcceptCtlSThree}{8/15} %
\newcommand{\SLVinReplayMainSThree}{0/45} %
\newcommand{\SLControlVinRangeSThree}{\ensuremath{+0.067} to \ensuremath{+0.073}} %
\newcommand{\SLControlPOneRangeSThree}{\ensuremath{-0.103} to \ensuremath{-0.068}} %
\newcommand{\SLControlRefreshedSpearmanRangeSThree}{0.132 to 0.439} %
\newcommand{\SLIsolationStepsSThree}{850} %
\newcommand{\SLSurveyLoopsSThree}{56} %
\newcommand{\SLSurveyCurveCompleteSThree}{2} %
\newcommand{\SLSurveyCurvePartialSThree}{9} %
\newcommand{\SLSurveyJudgeErrorQuantifiedSThree}{11} %
\newcommand{\SLSurveyBothAndGamedSThree}{0} %
\newcommand{\SLJudgeErrorGenOneRTwoSThree}{\ensuremath{+0.063}} %
\newcommand{\SLJudgeErrorGenOneRTwoCISThree}{[\ensuremath{+0.057}, \ensuremath{+0.069}]} %
\newcommand{\SLTransferRangeMainSThree}{\ensuremath{+0.007} to \ensuremath{+0.149}} %
\newcommand{\SLControlTransferRangeSThree}{\ensuremath{+0.138} to \ensuremath{+0.171}} %
\newcommand{\SLControlJudgeErrorRangeSThree}{\ensuremath{-0.409} to \ensuremath{-0.058}} %

\newcommand{\SLTasksSOne}{3} %
\newcommand{\SLEvalPerGenSOne}{1{,}000} %
\newcommand{\SLStartsInBandSOne}{0.661 / 0.746 / 0.491} %
\newcommand{\SLJudgeFPRStartSOne}{0.000 to 0.227} %
\newcommand{\SLJudgeFPREndSOne}{0.000 to 1.000} %
\newcommand{\SLFTwoHitRateSOne}{48/114} %
\newcommand{\SLFTwoPriorBandThroughoutSOne}{74/114} %
\newcommand{\SLBlindGenerationsSOne}{3} %
\newcommand{\SLClassifierDriftMaxSOne}{12/300} %
\newcommand{\SLClassifierDriftZeroOrOneSOne}{98/114} %
\newcommand{\SLIsolationPosthocSOne}{114/114} %
\newcommand{\SLPerturbedPoolPiOnePooledSOne}{\ensuremath{+0.075}} %
\newcommand{\SLPerturbedPoolPiThreeTargetShiftSOne}{\ensuremath{-0.106}} %
\newcommand{\SLPerturbedPoolPiThreePooledSOne}{\ensuremath{+0.026}} %
\newcommand{\SLPerturbedPoolPiFourTargetShiftSOne}{\ensuremath{-0.126}} %
\newcommand{\SLPerturbedPoolPiFourPooledSOne}{\ensuremath{+0.030}} %
\newcommand{\SLExtendedChainsSOne}{8} %
\newcommand{\SLExtensionGenerationsInLoopSOne}{24} %
\newcommand{\SLExtensionAcceptedSOne}{0} %
\newcommand{\SLGapPositiveChainsSOne}{5 of 9} %
\newcommand{\SLDistinctTrajectoriesSOne}{87.6\% of 102{,}600} %
\newcommand{\SLSeedOutcomeAgreementSOne}{93.1\% of 17{,}100} %
\newcommand{\SLKendallTauSOne}{1.00} %
\newcommand{\SLKendallTauAllPairsSOne}{0.50} %
\newcommand{\SLPairedOutcomeIdenticalSOne}{86.2\% of 4{,}384} %
\newcommand{\SLFooledAcceptedFailuresSOne}{65 of 196} %

\IfFileExists{sl_ladder_macros.tex}{
\newcommand{\SLLadderPrimaryGens}{38}
\newcommand{\SLLadderFooledA}{65}
\newcommand{\SLLadderFooledB}{196}
\newcommand{\SLLadderAgreeJZero}{35}
\newcommand{\SLLadderAgreeJTwo}{27}
\newcommand{\SLLadderAgreeJThree}{24}
\newcommand{\SLLadderAgreeJFour}{27}
\newcommand{\SLLadderAgreeJFive}{24}
\newcommand{\SLLadderBestVlmAgree}{27}
}{}
\IfFileExists{sl_s1_probe_auc_macros.tex}{\input{sl_s1_probe_auc_macros.tex}}{}
\IfFileExists{sl_s1_vlm_arm_macros.tex}{
\newcommand{\SLVLMGenerationsSOne}{15} %
\newcommand{\SLVLMCertificatesSOne}{3} %
\newcommand{\SLVLMSameAsPredicateSOne}{11 of 15} %
\newcommand{\SLCZeroSameAsPredicateVLMArmSOne}{13 of 15} %
\newcommand{\SLVLMResolvedSOne}{6 of 15} %
\newcommand{\SLPredicateResolvedVLMArmSOne}{8 of 15} %
\newcommand{\SLVLMMedianStopSOne}{300} %
\newcommand{\SLPredicateMedianStopVLMArmSOne}{210} %
\newcommand{\SLVLMPositiveRollbacksSOne}{5} %
\newcommand{\SLVLMJudgeAttributableSOne}{1} %
\newcommand{\SLVLMBudgetAttributableSOne}{4} %
\newcommand{\SLVLMFalseRollbacksSOne}{2} %
}{}
\newcommand{\SLLadderAgreeJSix}{26}
\newcommand{\SLLadderEpisodesPrimary}{22{,}800}
\newcommand{\SLLadderFPRJSix}{0.08}
\newcommand{\SLLadderFPRJZero}{0.11}
\newcommand{\SLLadderSecPerKJFour}{371}
\newcommand{\SLLadderSecPerKJSix}{722}
\newcommand{\SLLadderTPRJSix}{0.69}
\newcommand{\SLLadderTPRJZero}{1.00}
\newcommand{\SLPostAuditAPerTask}{300}
\newcommand{\SLPostAuditBPerTask}{350}
\newcommand{\SLPostAuditInLoopFinalAMean}{\ensuremath{+0.21}}
\newcommand{\SLPostAuditInLoopFinalBMean}{\ensuremath{+0.02}}
\newcommand{\SLPostAuditInLoopKeptStart}{1}
\newcommand{\SLPostAuditSealedFinalAMean}{\ensuremath{+0.18}}
\newcommand{\SLPostAuditSealedFinalBMean}{\ensuremath{+0.06}}
\newcommand{\SLPostAuditSealedKeptStart}{2}
\newcommand{\SLPostAuditWorstFinalB}{\ensuremath{-0.13}}
\newcommand{\SLPostBudgetInLoopBudget}{8}
\newcommand{\SLPostBudgetInLoopJudge}{6}
\newcommand{\SLPostBudgetInLoopN}{45}
\newcommand{\SLPostBudgetInLoopSample}{3}
\newcommand{\SLPostBudgetSealedBudget}{7}
\newcommand{\SLPostBudgetSealedJudge}{9}
\newcommand{\SLPostBudgetSealedN}{45}
\newcommand{\SLPostBudgetSealedSample}{4}
\newcommand{\SLPostCtlGains}{13}
\newcommand{\SLPostCtlLoopPairs}{60}
\newcommand{\SLPostCtlLosses}{61}
\newcommand{\SLPostCtlNearZero}{2}
\newcommand{\SLPostCtlPairs}{76}
\newcommand{\SLPostCtlSeqAccFramesRange}{2{,}107 to 15{,}177}
\newcommand{\SLPostCtlSeqFrames}{576{,}274}
\newcommand{\SLPostCtlSeqGainsAcc}{11}
\newcommand{\SLPostCtlSeqMissed}{\ensuremath{+0.016} and \ensuremath{+0.022}}
\newcommand{\SLPostCtlWeakPairs}{15}
\newcommand{\SLPostCtlWeakStarts}{6}
\newcommand{\SLPostSimCells}{216}
\newcommand{\SLPostSimDriveFA}{0}
\newcommand{\SLPostSimDriveFrameFA}{10}
\newcommand{\SLPostSimDriveN}{1{,}499}
\newcommand{\SLPostSimIidPooled}{0.68\%}
\newcommand{\SLPostSimIidRun}{10.3\%}
\newcommand{\SLPostSimIidWorst}{2.1\%}
\newcommand{\SLPostSimInterPooled}{0.23\%}
\newcommand{\SLPostSimInterRun}{5.7\%}
\newcommand{\SLPostSimInterWorst}{1.2\%}
\newcommand{\SLPostSimLossAcc}{0}
\newcommand{\SLPostSimLossRuns}{9{,}000}
\newcommand{\SLPostSimReplicates}{189{,}000}
\newcommand{\SLPostSimSmallGainAcc}{75\% to 99\%}
\newcommand{\SLPostSimUnifPooled}{0.66\%}
\newcommand{\SLPostSimUnifRun}{9.1\%}
\newcommand{\SLPostSimUnifWorst}{1.9\%}
\newcommand{\SLPostSimUnionBound}{25\%}
\newcommand{\SLPostTransferCtlBostonMed}{\ensuremath{-0.082}}
\newcommand{\SLPostTransferCtlMixMed}{\ensuremath{-0.042}}
\newcommand{\SLPostTransferCtlMixRange}{\ensuremath{-0.062} to \ensuremath{-0.034}}
\newcommand{\SLPostTransferCtlPThreeMed}{\ensuremath{-0.052}}
\newcommand{\SLPostTransferCtlPThreeRange}{\ensuremath{-0.074} to \ensuremath{-0.044}}
\newcommand{\SLPostTransferCtlShare}{80\%}
\newcommand{\SLPostTransferCtlVinMed}{\ensuremath{+0.070}}
\newcommand{\SLPostTransferMainBostonMed}{\ensuremath{-0.055}}
\newcommand{\SLPostTransferMainMixMed}{\ensuremath{-0.045}}
\newcommand{\SLPostTransferMainMixRange}{\ensuremath{-0.195} to \ensuremath{+0.004}}
\newcommand{\SLPostTransferMainN}{45}
\newcommand{\SLPostTransferMainNeg}{44}
\newcommand{\SLPostTransferMainPThreeMed}{\ensuremath{-0.058}}
\newcommand{\SLSeqDecidedMainSThree}{38}
\newcommand{\SLSeqFramesMainSThree}{365{,}662}
\newcommand{\SLSeqPoolFramesMainSThree}{750{,}825}
\newcommand{\SLSeqShareMainSThree}{49\%}
\newcommand{\SLSeqStopMedianMainSThree}{5{,}446}
\newcommand{\SLSeqUndecidedMainSThree}{7}

\IfFileExists{sl_vehicle_macros.tex}{
\newcommand{\SLVehicleNavsimILSFour}{0.861}
\newcommand{\SLVehicleNavsimRLSFour}{0.904}
\newcommand{\SLVehicleNavsimPairedSFour}{+0.043 (95\% interval +0.039 to +0.048)}
\newcommand{\SLVehicleNavsimTokensSFour}{12,118}
}{\input{sl_vehicle_pending_macros.tex}}
\newcommand{\SLVehicleSubDACIL}{0.946}
\newcommand{\SLVehicleSubDACOnlyIL}{440}
\newcommand{\SLVehicleSubDACOnlyRL}{72}
\newcommand{\SLVehicleSubDACRL}{0.976}
\newcommand{\SLVehicleSubDDCIL}{0.981}
\newcommand{\SLVehicleSubDDCRL}{0.967}
\newcommand{\SLVehicleSubEPIL}{0.804}
\newcommand{\SLVehicleSubEPRL}{0.858}
\newcommand{\SLVehicleSubNCOnlyIL}{101}
\newcommand{\SLVehicleSubNCOnlyRL}{88}
\newcommand{\SLVehicleSubScenes}{12{,}118}

\newcommand{\SLVehAurocJFour}{0.11}
\newcommand{\SLVehAurocJThree}{0.07}
\newcommand{\SLVehAurocJTwo}{0.17}
\newcommand{\SLVehClipLeft}{9.7}
\newcommand{\SLVehGuessLeft}{2 of 12}
\newcommand{\SLVehGuessRight}{2 of 2}
\newcommand{\SLVehGuessStraight}{4 of 4}
\newcommand{\SLVehGuessStraightDK}{2}
\newcommand{\SLVehLeftCS}{[\ensuremath{-0.50}, \ensuremath{+0.57}]}
\newcommand{\SLVehLeftCompleted}{0 of 12}
\newcommand{\SLVehLeftFailMax}{57}
\newcommand{\SLVehLeftFailMin}{30}
\newcommand{\SLVehLeftLength}{144}
\newcommand{\SLVehLeftMeanDiff}{\ensuremath{+0.072}}
\newcommand{\SLVehLeftMeanDiffM}{10.4}
\newcommand{\SLVehLeftMeanIL}{41.2}
\newcommand{\SLVehLeftMeanRL}{51.6}
\newcommand{\SLVehLeftPairs}{6}
\newcommand{\SLVehLeftRLFurther}{6 of 6}
\newcommand{\SLVehLeftSignP}{0.016}
\newcommand{\SLVehPlanPeriod}{0.80}
\newcommand{\SLVehPrefMaxDiff}{0.02}
\newcommand{\SLVehRightLength}{143}
\newcommand{\SLVehRightMeanIL}{83.1}
\newcommand{\SLVehRightMeanRL}{98.2}
\newcommand{\SLVehRightPairs}{1}
\newcommand{\SLVehScored}{20}
\newcommand{\SLVehStraightCompleted}{6 of 6}
\newcommand{\SLVehStraightLength}{60}
\newcommand{\SLVehStraightPairs}{3}

\renewcommand{\SLStartsInBand}{\SLStartsInBandSOne}
\renewcommand{\SLEvalPerGen}{\SLEvalPerGenSOne}
\newcommand{\SLEvalPerGenInLoopSOne}{300}
\newcommand{\SLCollectPerGenSOne}{300}
\newcommand{\SLPoolFramesSThree}{16{,}685}
\newcommand{\SLInLoopFramesSThree}{5{,}000}
\newcommand{\SLImproverFramesSThree}{20{,}000}
\newcommand{\SLProposalsSThree}{20}
\newcommand{\SLAlpha}{0.05}
\newcommand{\SLMinN}{10}
\newcommand{\SLGapThresholdSThree}{0.02 PDMS}
\newcommand{\SLGapThresholdSOne}{0.10 in success rate}
\newcommand{\SLTeaserGensSOne}{90}
\newcommand{\SLTeaserOverSOne}{5}
\newcommand{\SLTeaserGensSThree}{45}
\newcommand{\SLTeaserOverSThree}{10}

\newcommand{\SLJudgesHardN}{196}
\newcommand{\SLJudgesBlindN}{36}
\newcommand{\SLJudgesHardJZero}{33\%}
\newcommand{\SLJudgesBlindJZero}{100\%}
\newcommand{\SLJudgesHardJTwo}{69\%}
\newcommand{\SLJudgesHardJThree}{23\%}
\newcommand{\SLJudgesHardJFour}{34\%}
\newcommand{\SLJudgesHardJFive}{42\%}
\newcommand{\SLJudgesHardJSix}{3\%}
\newcommand{\SLJudgesBlindJSix}{0\%}
\newcommand{\SLJudgesHardFewer}{two}
\newcommand{\SLJudgesHardMore}{three}
\newcommand{\SLJudgesOthers}{five}
\newcommand{\SLJudgesSealedGens}{31}
\newcommand{\SLJudgesSealedAgreeJZero}{25}
\newcommand{\SLJudgesSealedAgreeBestOther}{23}

\newcommand{\SLFairBootFiftyGains}{10.8}
\newcommand{\SLFairBootFiftyNullFA}{0.8\%}
\newcommand{\SLFairBootFiftyShare}{22\%}
\newcommand{\SLFairCSFiftyGains}{1.1}
\newcommand{\SLFairGains}{12}
\newcommand{\SLFairLosses}{59}
\newcommand{\SLFairNearZero}{5}
\newcommand{\SLFairNullPools}{18{,}972}
\newcommand{\SLFairOrders}{500}
\newcommand{\SLFairPairs}{76}
\newcommand{\SLFairPeekEvery}{10}
\newcommand{\SLFairPeekNearAcc}{1.16}
\newcommand{\SLFairPeekNearMaxPair}{88\%}
\newcommand{\SLFairPeekNullFA}{5.9\%}
\newcommand{\SLFairPeekNullFAMax}{11.4\%}
\newcommand{\SLFairSeqGains}{11.2}
\newcommand{\SLFairSeqOverBootFifty}{2.4}
\newcommand{\SLFairSeqShare}{53\%}

\newcommand{\SLAddOracleFrames}{20{,}000}
\newcommand{\SLAddOracleTargetPDMS}{0.965}
\newcommand{\SLAddOraclePolicyPDMS}{0.887}
\newcommand{\SLAddOracleGainMid}{${-}0.008$}
\newcommand{\SLAddOracleCIMid}{[${-}0.012$, ${-}0.004$]}
\newcommand{\SLAddOracleGainLow}{${-}0.018$}
\newcommand{\SLAddOracleCILow}{[${-}0.021$, ${-}0.014$]}
\newcommand{\SLAddOracleGainHigh}{${-}0.108$}
\newcommand{\SLAddOracleCIHigh}{[${-}0.117$, ${-}0.099$]}
\newcommand{\SLAddOracleBostonFrames}{16{,}685}
\newcommand{\SLAddCalibSteps}{2{,}000}
\newcommand{\SLAddCalibOwnMin}{${-}0.30$}
\newcommand{\SLAddCalibOwnMax}{${-}0.13$}
\newcommand{\SLAddCalibOwnLarge}{${-}0.27$}
\newcommand{\SLAddCalibDemo}{$+0.31$}
\newcommand{\SLAddCalibMix}{$+0.06$}
\newcommand{\SLAddCalibMixCI}{[${-}0.06$, $+0.17$]}
\newcommand{\SLAddEvalDisagreeZeroOrOne}{69/114}
\newcommand{\SLAddEvalDisagreeMax}{68/300}
\newcommand{\SLAddFlips}{7}
\newcommand{\SLAddFlipsOf}{114}
\newcommand{\SLAddFlipsClassifierRuns}{4}
\newcommand{\SLAddFlipsLenient}{5}
\newcommand{\SLAddWarnManipPairs}{72}
\newcommand{\SLAddWarnManipPos}{13}
\newcommand{\SLAddWarnManipFeatures}{48}
\newcommand{\SLAddWarnManipAbove}{20}
\newcommand{\SLAddWarnManipTop}{0.81}
\newcommand{\SLAddWarnManipTopCI}{[0.69, 0.91]}
\newcommand{\SLAddWarnManipTopWithin}{0.80}
\newcommand{\SLAddWarnManipToggles}{0.68}
\newcommand{\SLAddWarnManipTogglesWithin}{0.31}
\newcommand{\SLAddWarnDrivePairs}{36}
\newcommand{\SLAddWarnDrivePos}{17}
\newcommand{\SLAddWarnDriveFeatures}{13}
\newcommand{\SLAddWarnDriveAbove}{2}
\newcommand{\SLAddWarnDriveTop}{0.67}
\newcommand{\SLAddNeedSealedManipN}{29}
\newcommand{\SLAddNeedSealedManipMedian}{4.7}
\newcommand{\SLAddNeedClassManipN}{48}
\newcommand{\SLAddNeedClassManipMedian}{7.4}
\newcommand{\SLAddNeedSealedDriveN}{8}
\newcommand{\SLAddNeedSealedDriveMedian}{5.6}
\newcommand{\SLAddNeedSealedDriveMax}{242}
\newcommand{\SLAddNeedManipReadings}{77}
\newcommand{\SLAddNeedManipMin}{133}
\newcommand{\SLAddNeedManipMax}{48{,}024}
\newcommand{\SLAddDetN}{300}
\newcommand{\SLAddDetTPR}{0.82}
\newcommand{\SLAddDetFPR}{0.24}
\newcommand{\SLAddDetAUC}{0.85}
\newcommand{\SLAddDetJ}{0.585}
\newcommand{\SLAddDetFit}{510}
\newcommand{\SLAddDetJTaskOne}{0.47}
\newcommand{\SLAddDetJTaskTwo}{0.36}
\newcommand{\SLAddDetJTaskThree}{0.94}

\newcommand{\SLSubsetConfigs}{six}
\newcommand{\SLSubsetStartPDMS}{0.698}
\newcommand{\SLSubsetSelFrames}{3{,}731}
\newcommand{\SLSubsetSelShare}{22\%}
\newcommand{\SLSubsetFrames}{12{,}954}
\newcommand{\SLSubsetLogs}{229}
\newcommand{\SLSubsetCandidates}{45}
\newcommand{\SLSubsetNegative}{45}
\newcommand{\SLSubsetBelowZero}{43}
\newcommand{\SLSubsetKeptA}{${-}0.028$}
\newcommand{\SLSubsetKeptB}{${-}0.012$}
\newcommand{\SLSubsetKeptC}{${-}0.026$}

\providecommand{\SLJudgeGapGenOneInLoopROneSThree}{}\renewcommand{\SLJudgeGapGenOneInLoopROneSThree}{\ensuremath{+0.045}}
\providecommand{\SLJudgeGapGenOneCIInLoopROneSThree}{}\renewcommand{\SLJudgeGapGenOneCIInLoopROneSThree}{[\ensuremath{+0.037}, \ensuremath{+0.052}]}
\providecommand{\SLJudgeGapGenOneInLoopRTwoSThree}{}\renewcommand{\SLJudgeGapGenOneInLoopRTwoSThree}{\ensuremath{+0.095}}
\providecommand{\SLJudgeGapGenOneCIInLoopRTwoSThree}{}\renewcommand{\SLJudgeGapGenOneCIInLoopRTwoSThree}{[\ensuremath{+0.087}, \ensuremath{+0.104}]}
\providecommand{\SLJudgeErrorGenOneRTwoSThree}{}\renewcommand{\SLJudgeErrorGenOneRTwoSThree}{\ensuremath{+0.070}}
\providecommand{\SLJudgeErrorGenOneRTwoCISThree}{}\renewcommand{\SLJudgeErrorGenOneRTwoCISThree}{[\ensuremath{+0.061}, \ensuremath{+0.079}]}
\providecommand{\SLTransferRangeMainSThree}{}\renewcommand{\SLTransferRangeMainSThree}{\ensuremath{+0.008} to \ensuremath{+0.123}}
\providecommand{\SLSealedGainFirstMeanSThree}{}\renewcommand{\SLSealedGainFirstMeanSThree}{\ensuremath{-0.028}}
\providecommand{\SLControlVinRangeSThree}{}\renewcommand{\SLControlVinRangeSThree}{\ensuremath{+0.047} to \ensuremath{+0.058}}
\providecommand{\SLControlPOneRangeSThree}{}\renewcommand{\SLControlPOneRangeSThree}{\ensuremath{-0.107} to \ensuremath{-0.075}}
\providecommand{\SLControlTransferRangeSThree}{}\renewcommand{\SLControlTransferRangeSThree}{\ensuremath{+0.125} to \ensuremath{+0.160}}
\providecommand{\SLControlJudgeErrorRangeSThree}{}\renewcommand{\SLControlJudgeErrorRangeSThree}{\ensuremath{-0.376} to \ensuremath{-0.034}}
\providecommand{\SLPostCtlGains}{}\renewcommand{\SLPostCtlGains}{12}
\providecommand{\SLPostCtlLosses}{}\renewcommand{\SLPostCtlLosses}{59}
\providecommand{\SLPostCtlNearZero}{}\renewcommand{\SLPostCtlNearZero}{5}
\providecommand{\SLPostCtlSeqGainsAcc}{}\renewcommand{\SLPostCtlSeqGainsAcc}{11}
\providecommand{\SLPostCtlSeqMissed}{}\renewcommand{\SLPostCtlSeqMissed}{\ensuremath{+0.019}}
\providecommand{\SLPostCtlSeqFrames}{}\renewcommand{\SLPostCtlSeqFrames}{561{,}665}
\providecommand{\SLPostCtlSeqAccFramesRange}{}\renewcommand{\SLPostCtlSeqAccFramesRange}{2{,}098 to 14{,}452}
\providecommand{\SLSeqDecidedMainSThree}{}\renewcommand{\SLSeqDecidedMainSThree}{38}
\providecommand{\SLSeqUndecidedMainSThree}{}\renewcommand{\SLSeqUndecidedMainSThree}{7}
\providecommand{\SLSeqFramesMainSThree}{}\renewcommand{\SLSeqFramesMainSThree}{354{,}754}
\providecommand{\SLSeqShareMainSThree}{}\renewcommand{\SLSeqShareMainSThree}{47\%}
\providecommand{\SLSeqStopMedianMainSThree}{}\renewcommand{\SLSeqStopMedianMainSThree}{5{,}316}
\providecommand{\SLPostTransferMainN}{}\renewcommand{\SLPostTransferMainN}{45}
\providecommand{\SLPostTransferMainNeg}{}\renewcommand{\SLPostTransferMainNeg}{44}
\providecommand{\SLPostTransferMainPThreeMed}{}\renewcommand{\SLPostTransferMainPThreeMed}{\ensuremath{-0.054}}
\providecommand{\SLPostTransferMainMixMed}{}\renewcommand{\SLPostTransferMainMixMed}{\ensuremath{-0.043}}
\providecommand{\SLPostTransferMainMixRange}{}\renewcommand{\SLPostTransferMainMixRange}{\ensuremath{-0.186} to \ensuremath{+0.006}}
\providecommand{\SLPostTransferMainBostonMed}{}\renewcommand{\SLPostTransferMainBostonMed}{\ensuremath{-0.054}}
\providecommand{\SLPostTransferCtlShare}{}\renewcommand{\SLPostTransferCtlShare}{70\%}
\providecommand{\SLPostTransferCtlVinMed}{}\renewcommand{\SLPostTransferCtlVinMed}{\ensuremath{+0.053}}
\providecommand{\SLPostTransferCtlPThreeMed}{}\renewcommand{\SLPostTransferCtlPThreeMed}{\ensuremath{-0.045}}
\providecommand{\SLPostTransferCtlPThreeRange}{}\renewcommand{\SLPostTransferCtlPThreeRange}{\ensuremath{-0.064} to \ensuremath{-0.038}}
\providecommand{\SLPostTransferCtlMixMed}{}\renewcommand{\SLPostTransferCtlMixMed}{\ensuremath{-0.037}}
\providecommand{\SLPostTransferCtlMixRange}{}\renewcommand{\SLPostTransferCtlMixRange}{\ensuremath{-0.053} to \ensuremath{-0.029}}
\providecommand{\SLPostTransferCtlBostonMed}{}\renewcommand{\SLPostTransferCtlBostonMed}{\ensuremath{-0.089}}
\providecommand{\SLSubsetNegative}{}\renewcommand{\SLSubsetNegative}{43}
\providecommand{\SLSubsetBelowZero}{}\renewcommand{\SLSubsetBelowZero}{42}
\providecommand{\SLSubsetSame}{}\renewcommand{\SLSubsetSame}{43}
\providecommand{\SLSubsetChanged}{}\renewcommand{\SLSubsetChanged}{2}
\providecommand{\SLSubsetKeptA}{}\renewcommand{\SLSubsetKeptA}{${-}0.026$}
\providecommand{\SLSubsetKeptB}{}\renewcommand{\SLSubsetKeptB}{${-}0.012$}
\providecommand{\SLSubsetKeptC}{}\renewcommand{\SLSubsetKeptC}{${-}0.023$}

\newcommand{\SLWorkSeeds}{3}
\newcommand{\SLWorkGens}{5}
\newcommand{\SLWorkCands}{32}
\newcommand{\SLWorkAccGainLo}{\ensuremath{+0.029}}
\newcommand{\SLWorkAccGainHi}{\ensuremath{+0.043}}
\newcommand{\SLWorkUndGainLo}{\ensuremath{-0.004}}
\newcommand{\SLWorkUndGainHi}{\ensuremath{+0.017}}
\newcommand{\SLWorkStartPDMS}{0.638}
\newcommand{\SLWorkEndPDMS}{0.674}
\newcommand{\SLWorkVinEndPDMS}{0.689}
\newcommand{\SLWorkCriticEndPDMS}{0.649}
\newcommand{\SLWorkCriticStuckSeeds}{two}
\newcommand{\SLWorkVinAcc}{23}
\newcommand{\SLWorkSeqShare}{85\%}
\newcommand{\SLWorkFirst}{11}
\newcommand{\SLWorkLater}{21}
\newcommand{\SLWorkFirstCriticAcc}{one}
\newcommand{\SLWorkLaterVinAcc}{12}
\newcommand{\SLWorkSealedRise}{0.036}
\newcommand{\SLWorkVinRise}{0.051}

\newcommand{\SLPredfixChains}{9}
\newcommand{\SLPredfixGens}{45}
\newcommand{\SLPredfixKept}{13}
\newcommand{\SLPredfixKeptCertGain}{13}
\newcommand{\SLPredfixWorseKept}{0}
\newcommand{\SLPredfixGainsLost}{1}
\newcommand{\SLPredfixSameAsSealed}{44}

\newcommand{\SLPThreeReplayN}{60}
\newcommand{\SLPThreeReplayRej}{20}

\newcommand{\SLSuppSurveyManipN}{24}
\newcommand{\SLSuppSurveyDriveN}{32}
\newcommand{\SLSuppManipGens}{114}
\newcommand{\SLSuppManipDiffer}{26}
\newcommand{\SLSuppManipDifferFive}{22}
\newcommand{\SLSuppManipAccRej}{1}
\newcommand{\SLSuppManipUndDec}{25}
\newcommand{\SLSuppManipMedianPairs}{191}
\newcommand{\SLSuppManipAtBudget}{32}
\newcommand{\SLSuppFprFirst}{0.05}
\newcommand{\SLSuppFprPeak}{0.32}
\newcommand{\SLSuppFprPeakGen}{6}
\newcommand{\SLSuppFprVlmLo}{0.28}
\newcommand{\SLSuppFprVlmHi}{0.46}
\newcommand{\SLSuppFprAbove}{31}
\newcommand{\SLSuppFprGens}{114}
\newcommand{\SLSuppLookScoreA}{0.73}
\newcommand{\SLSuppLookScoreB}{0.88}
\newcommand{\SLSuppLookPreA}{0.96}
\newcommand{\SLSuppLookPreB}{0.95}
\newcommand{\SLSuppAuditStepSealedN}{12}
\newcommand{\SLSuppAuditStepInLoopN}{13}
\newcommand{\SLSuppDriveMainGens}{45}
\newcommand{\SLSuppDriveCtlGens}{15}
\newcommand{\SLSuppDriveErrMain}{$-$0.05}
\newcommand{\SLSuppDriveErrCtl}{$-$0.18}
\newcommand{\SLSuppDriveGapMain}{+0.04}
\newcommand{\SLSuppDriveGapCtl}{+0.14}
\newcommand{\SLSuppDriveGapPos}{60}
\newcommand{\SLSuppDriveCriticAcc}{3}
\newcommand{\SLSuppDriveSealedAcc}{0}
\newcommand{\SLSuppDriveSealedRej}{38}
\newcommand{\SLSuppDriveSealedUnd}{7}
\newcommand{\SLSuppDriveCtlCriticRej}{13}
\newcommand{\SLSuppDriveCtlSealedRej}{1}
\newcommand{\SLSuppDriveCtlSealedUnd}{14}
\newcommand{\SLSuppCityVegas}{$-$0.033}
\newcommand{\SLSuppCityPitt}{$-$0.100}
\newcommand{\SLSuppCitySing}{$-$0.053}
\newcommand{\SLSuppCityBoston}{$-$0.054}
\newcommand{\SLSuppTransferFrames}{8{,}415}
\newcommand{\SLSuppRulesPairs}{76}

\newcommand{\SLExtraNullLow}{${-}0.001$}
\newcommand{\SLExtraNullMid}{$+0.000$}
\newcommand{\SLExtraNullHigh}{${-}0.002$}
\newcommand{\SLExtraNullFrames}{5{,}000}
\newcommand{\SLExtraReplayWindows}{893}
\newcommand{\SLExtraReplayDiff}{0.68}
\newcommand{\SLExtraReplaySpreadIL}{0.11}
\newcommand{\SLExtraReplaySpreadRL}{0.12}
\newcommand{\SLExtraReplayShare}{85\%}
\newcommand{\SLExtraCamBlack}{${-}0.126$}
\newcommand{\SLExtraCamGrey}{${-}0.107$}
\newcommand{\SLExtraCamFrames}{3{,}731}
\newcommand{\SLExtraDemoPath}{$+0.25$}
\newcommand{\SLExtraDemoPathCI}{[$+0.15$, $+0.35$]}
\newcommand{\SLExtraDemoPathN}{120}

\newcommand{\SLWorReplayManipSame}{52}
\newcommand{\SLWorReplayDriveSame}{50}

\newcommand{\SLFixedGens}{114}
\newcommand{\SLFixedSame}{109}
\newcommand{\SLFixedAccepted}{25}

\title{Keep or Roll Back?\\Out-of-Loop Judges for Self-Improving VLA and End-to-End Embodied Policies}

\author{Zilin Huang$^{1}$\qquad Zhengyang Wan$^{1}$\qquad Kangan Qian$^{2}$\qquad Zihao Sheng$^{1}$\qquad Sikai Chen$^{1,}$\thanks{Corresponding author: \texttt{sikai.chen@wisc.edu}}\\[2pt]
$^{1}$University of Wisconsin--Madison\qquad $^{2}$University of California, Santa Cruz\\[2pt]
{\small Project page: \url{https://zilin-huang.github.io/SealedLoop/}}}

\newtheorem{supplemma}{Lemma}

\newtheorem{suppcorollary}{Corollary}

\newtheorem{supproposition}{Proposition}

\begin{document}
\maketitle
\begin{strip}
\centering
\includegraphics[width=\textwidth]{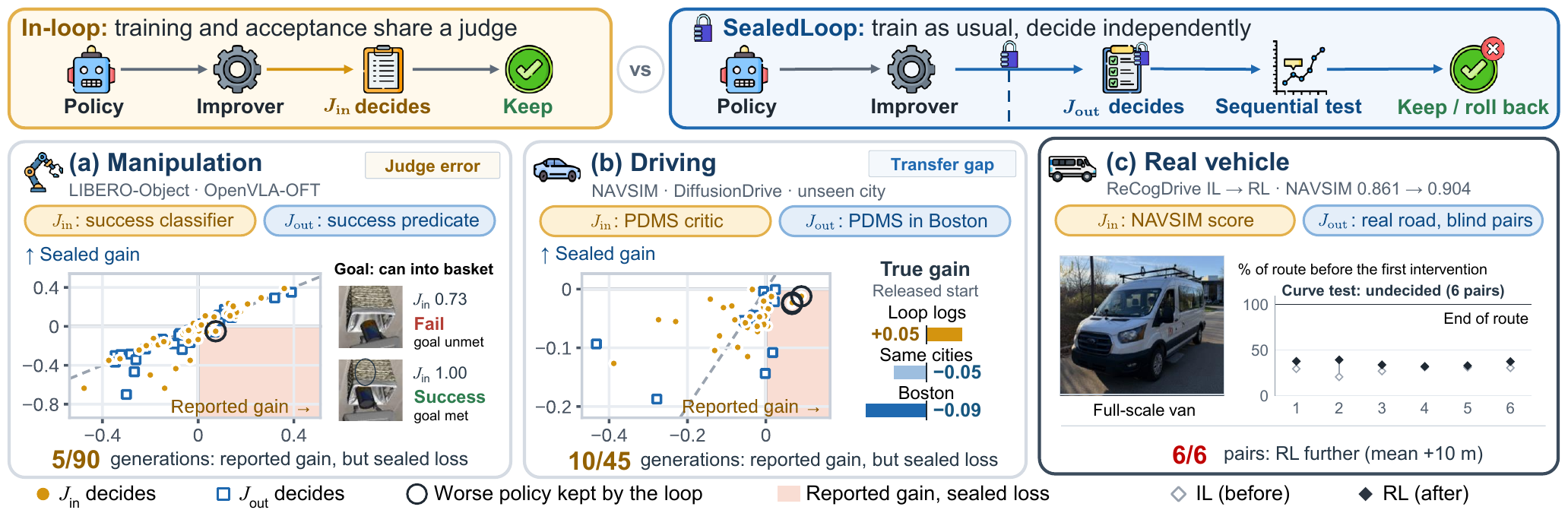}
\captionof{figure}{\textbf{Self-improvement loops report gains that a sealed judge does not find.} Top: today the loop's own judge decides what is kept; \method lets a sealed judge decide through a sequential test. (a, b) Reported gain against sealed gain, one point per generation; the shaded quadrant holds reported gains with a sealed loss. (c) Real-vehicle field experiment on a published improvement step, in blind pairs.}
\label{fig:teaser}
\end{strip}

\begin{abstract}
Self-improvement is now a practical recipe for embodied policies: a policy collects experience, is retrained, and is kept when a judge rates it higher. However, the judge is usually trained or chosen inside the same loop, so its evidence is not independent of the policy. We introduce \method, an evaluation framework that adds a \emph{sealed judge}: an evaluator outside the loop that the improver cannot see or optimize against. A new policy is kept only when a sequential test on the sealed judge confirms the gain, with a bounded probability of keeping a worse one. Running the same loop under each judge, we separate two sources of a false gain, which we call \emph{judge error} (the loop's judge is wrong on its own data) and \emph{transfer gap} (a real gain there does not carry over to unseen conditions). We evaluate on LIBERO manipulation, NAVSIM driving, and a full-scale real vehicle. In manipulation, a success classifier accepted \SLFooledAcceptedFailuresSOne{} failures in one generation. In driving, policies that gain on the loop's own frames lose in an unseen city, so even an exact in-loop judge would keep \SLVinReplayBlockAcceptCtlSThree{} of them. When the sealed test decides, a driving loop built to improve ends with a better policy than under its own critic (\SLWorkEndPDMS{} against \SLWorkCriticEndPDMS{} PDMS, from \SLWorkStartPDMS{}), and on driving pairs with known sealed gain the test accepted \SLPostCtlSeqGainsAcc{} of \SLPostCtlGains{} real gains and none of \SLPostCtlLosses{} losses. In a real-world field experiment on a published improvement step, the reinforced version drove further into a curve than the version it started from in all \SLVehLeftPairs{} blind pairs, though neither completed it. These results show that evidence from inside the loop, however accurate, is not enough to decide whether a policy has improved. \method offers a foundation for trustworthy self-improvement of embodied policies, in which progress is judged by evidence the improving system cannot influence.
\end{abstract}

\section{Introduction}
\label{sec:intro}

Self-improvement is now a practical recipe for embodied policies: a policy rolls out, a judge labels its episodes, the policy is fine-tuned on what was judged good, and the cycle repeats. Robot policies improve this way through reinforcement-learning fine-tuning and deployment experience~\cite{ax2509_09674,ax2510_25889,ax2410_21845,ax2502_05450,recap2025,ax2509_15937,ax2606_10305}, and driving planners through self-play and reinforcement learning against a simulator's score~\cite{ax2502_03349,ax2606_19641,ax2506_08052}. Every round ends with the same decision: is the new policy better, and should it replace the old one? Run without a human, the cycle is a bounded form of recursive self-improvement~\cite{ax2607_07663,axcs_0309048,ax2410_04444,ax2608_24735}, and it is only as reliable as this decision. In most loops, however, the judge that supplies the training signal or selects the training episodes also provides the evidence for this decision. The evidence is therefore not independent of the policy it evaluates. In language-model self-improvement, the loops that run reliably without a human are those whose judge the optimizer cannot influence, such as a code executor or a test suite~\cite{ax2505_22954,ax2506_13131,ax2505_03335}, whereas loops with a learned judge stop improving within a few rounds~\cite{ax2412_02674,ax2507_00075}.

Embodied loops face the same problem. The gain a loop reports can disagree with an independent evaluation for two reasons. First, the judge may mismeasure performance on the conditions it observes. Learned judges carry measurement error even before a policy is optimized against them: the best of 22 general robot reward models differs from human-verified progress labels by 0.67 points on a five-point scale on average~\cite{ax2601_00675}. Optimization then widens the error. A policy trained on a learned reward overshoots the true objective~\cite{gao2022overopt,ax2505_18126,ax2606_21090,ax2606_14629}, and progress rewards can be gamed by stopping short of the goal~\cite{ax2512_23703}. Second, a genuine gain on those conditions may not transfer to conditions the loop has not seen. We call the first \emph{judge error} and the second the \emph{transfer gap}. \Cref{fig:teaser} shows one case of each. In manipulation, a success classifier accepts failed placements that look like successes (a). In driving, gains on the loop's own evaluation frames turn into losses in an unseen city (b). A more accurate judge reduces judge error. The transfer gap can only be seen with evidence from outside the loop. But physical evaluation is slow and cannot be replayed, so embodied loops rarely collect such evidence.

A wrong decision does not stay in its round, because each kept policy collects the data for the next one. In driving, trajectory refinement guided by a learned critic peaks after two to three iterations~\cite{ax2508_09158}. Seeing such effects requires following a loop generation by generation and counting the cases where the policy exploits its judge. Among the \SLSurveyLoopsSThree{} embodied loops we surveyed, a per-generation curve with a quantified case of such exploitation has not yet been reported (\cref{tab:prior}). How the two sources affect successive acceptance decisions has not been measured. Measuring it requires an evaluator the improver cannot influence and an acceptance rule with a controlled error rate. Both exist elsewhere: sealed evaluation and anytime-valid acceptance for software agents~\cite{seal2026,ax2608_12841,ax2607_00871}, and sequential tests for comparing robot policies~\cite{ax2603_13616}. We bring them to embodied self-improvement, where the judge is learned from the loop's own data and sealed conditions differ from those the loop sees.

We propose \method, an evaluation framework that adds a sealed judge to the loop (\cref{fig:method}). A \emph{sealed judge} is an evaluator outside the loop whose data, parameters and schedule are hidden from the improver. It runs beside the loop, and a new policy is kept only when an anytime-valid sequential test~\cite{waudbysmith2024betting} on it certifies the gain. The sealed conditions are fixed before the improver runs and drawn independently of it. The probability that the test keeps a worse policy therefore does not exceed a level set in advance, whatever the improver does. The test is also sequential, so each decision comes with the amount of evaluation it used. Running the same loop twice, once with its own judge deciding and once with the sealed judge deciding, shows what the choice of the deciding judge changes. Both judges score every generation in either run. The difference between the gain the loop reports and the sealed gain, the \emph{judge gap}, can therefore be measured in every generation. The true score on the loop's own data separates it into judge error and transfer gap.

We apply \method to two simulated loops, LIBERO manipulation and NAVSIM driving, and to a full-scale real vehicle. In manipulation the judge gap is judge error: in one generation the success classifier accepted \SLFooledAcceptedFailuresSOne{} failures of the policy it trained, and \SLTeaserOverSOne{} of \SLTeaserGensSOne{} generations report a gain where the sealed judge finds a loss (\cref{fig:teaser}a). In driving a transfer gap appears in every generation, and \SLTeaserOverSThree{} of \SLTeaserGensSThree{} report a gain with a sealed loss (\cref{fig:teaser}b). Policies started from the released checkpoint gain on the loop's own evaluation frames and lose in an unseen city, so even an exact in-loop judge, one with no error on the loop's own data, would keep \SLVinReplayBlockAcceptCtlSThree{} of them. The sealed test still accepts real improvements: on driving pairs whose sealed gain is known, it accepted \SLPostCtlSeqGainsAcc{} of \SLPostCtlGains{} real gains and none of \SLPostCtlLosses{} losses.
In a driving loop built to improve, it certified all \SLWorkFirst{} first-step gains (\SLWorkAccGainLo{} to \SLWorkAccGainHi{} PDMS), of which the loop's learned critic accepted \SLWorkFirstCriticAcc{}.
In a real-world field experiment, the reinforced version of ReCogDrive~\cite{ax2506_08052}, whose gain is measured in simulation, drove further into a left curve in all \SLVehLeftPairs{} blind pairs, but neither completed it (\cref{fig:teaser}c).

Our contributions are:
\begin{itemize}
\item \textbf{We propose \method, an evaluation framework for embodied self-improvement loops.} A judge sealed from the improver, rather than the judge the loop trains with, decides each update through a sequential test, and running the loop under both judges separates a false gain into judge error and transfer gap.
\item \textbf{We show that evidence from inside the loop is not enough to approve an update.} On LIBERO and NAVSIM, in-loop judges keep policies that are not better, and even an exact in-loop judge would keep \SLVinReplayBlockAcceptCtlSThree{} of the driving policies that lose in an unseen city.
\item \textbf{We test a published improvement step on a full-scale real vehicle with a sealed judge.} In a blind, paired field experiment, the direction of its simulator gain holds on the road: the reinforced version goes further into a curve in all \SLVehLeftPairs{} pairs, though neither version completes it.
\end{itemize}
\section{Related Work}
\label{sec:related}

\begin{table}[t]
\centering\footnotesize
\setlength{\tabcolsep}{2.5pt}
\begin{tabular}{@{}>{\columncolor{white}[0pt][\tabcolsep]}llll>{\columncolor{white}[\tabcolsep][0pt]}l@{}}
\toprule
Loop & Judge & Rounds & Curve & Exploitation \\
\midrule
\rowcolor{black!8}\multicolumn{5}{@{}>{\columncolor{white}[0pt][0pt]}l@{}}{\emph{Manipulation}} \\
SimpleVLA-RL~\cite{ax2509_09674} & simulator & n.s. & no & no \\
HIL-SERL~\cite{ax2410_21845} & classifier + human & n.s. & no & no \\
$\pi^{*}_{0.6}$ (RECAP)~\cite{recap2025} & human raters & 2 & partial & no \\
VLAC~\cite{ax2509_15937} & VLM critic & n.s. & no & no \\
Robo-Dopamine~\cite{ax2512_23703} & reward model & n.s. & no & described \\
SARM2~\cite{ax2606_10305} & reward model & 3 & partial & mentioned \\
VLAW~\cite{ax2602_12063} & VLM reward model & 2 & partial & error only \\
\midrule
\rowcolor{black!8}\multicolumn{5}{@{}>{\columncolor{white}[0pt][0pt]}l@{}}{\emph{Driving}} \\
GIGAFLOW~\cite{ax2502_03349} & simulator rules & 1 & no & error only \\
ReCogDrive~\cite{ax2506_08052} & simulator score & 1 & no & no \\
EvaDrive~\cite{ax2508_09158} & learned critic & $\geq$4$^\dagger$ & partial & no \\
R$^2$LPL~\cite{ax2606_30537} & rule detectors & 5 & \good yes & no \\
AD-R1~\cite{ax2511_20325} & world-model critic & 1 & no & error only \\
\midrule
\ours This paper (both) & in-loop vs.\ sealed & 5 to 8 & \good yes & \good quantified \\
\bottomrule
\end{tabular}

\caption{\textbf{Embodied self-improvement loops and their judges}, a selection of the \SLSurveyLoopsSThree{} surveyed (all in \supp{Survey}). Rounds: improvement rounds reported (n.s.: not stated; $^\dagger$refinement iterations at inference). Curve: results per generation. Exploitation: policy gaming its judge is quantified (error only: the judge's error).}
\label{tab:prior}
\end{table}

\rhead{Self-improvement and its verifiers.}
Self-improvement loops run without a human when the optimizer cannot influence their judge, as with agents judged by an executor or a test suite~\cite{ax2505_22954,ax2510_21614,ax2504_15228,ax2505_03335,ax2203_14465,ax2603_25681,ax2607_07663}. Loops that update weights under a learned judge plateau within a few rounds~\cite{ax2401_10020,ax2412_02674,ax2507_00075}: judges favor outputs like their own~\cite{ax2410_21819}, and learned rewards are over-optimized~\cite{gao2022overopt,ax2505_18126}. For software agents, the judge has been moved out of the loop by hidden-rollout harnesses~\cite{seal2026}, test windows scored once~\cite{ax2608_12841}, error-controlled verifiers~\cite{ax2512_03109,ax2607_00871} and pre-registered analyses~\cite{ax2606_31511}. In embodied loops the judge is learned from the loop's own data and the sealed conditions differ from those the loop sees, so a reported gain can fail through judge error or a transfer gap; \method measures both, where a sealed evaluation alone gives only their sum. Anytime-valid inference~\cite{ramdas2023gametheoretic,waudbysmith2024betting} has been used to compare robot policies~\cite{ax2603_13616,ax2405_05439}, evaluate language models with fewer samples~\cite{ax2606_20820} and tell a drifting judge from a drifting system~\cite{ax2606_15474}. We use it as the loop's acceptance rule, bounding the probability of keeping a worse policy.

\rhead{Self-improving embodied policies.}
Reinforcement-learning fine-tuning of vision-language-action models is judged by the simulator~\cite{ax2509_09674,ax2510_25889,ax2505_17016}. On hardware, the judge is a human, a reward classifier trained in the loop~\cite{ax2410_21845,ax2502_05450,ax2412_09858,recap2025,ax2606_10305}, a vision-language critic~\cite{ax2509_15937}, a synthetic verifier~\cite{ax2606_19980} or a world model~\cite{ax2505_12705,ax2511_09515,ax2602_12063,ax2602_11075}. Such judges carry measurement error before any optimization~\cite{ax2601_00675,ax2511_20325}, and exploitation has so far been described qualitatively~\cite{ax2512_23703,ax2606_10305}. Driving planners trained against rule scores~\cite{ax2502_03349,ax2502_14706,ax2606_19641,ax2506_08052,ax2512_07745,ax2509_17940} report single runs, and learned-critic refinement peaks or saturates within a few iterations~\cite{ax2508_09158,ax2604_27366}. Across the \SLSurveyLoopsSThree{} surveyed loops, a per-generation curve with a quantified case of exploitation has not yet been reported (\cref{tab:prior}). Transfer is a further open problem: open-loop and closed-loop metrics disagree~\cite{ax2312_03031,navsim2024,ax2506_04218}, a benchmark gain can reverse when only the executor changes~\cite{ax2609_34440}, and driving models lose 22.8\% of their success rate under distribution shift~\cite{ax2604_08535}, while reinforcement-learning fine-tuning leaves visual robustness unchanged~\cite{ax2505_19789}. Simulated evaluation must be calibrated against hardware~\cite{ax2405_05941,ax2510_04354}. Physical evidence is rare: one study compares two policies over about 200\,km in three runs~\cite{ax2606_19836}. Building on these loops, we measure judge error and the transfer gap generation by generation on two embodiments, and compare both versions of a published improvement step in blind pairs on a full-scale vehicle.

\section{Self-Improvement under a Sealed Judge}
\label{sec:method}

\definecolor{slin}{HTML}{D55E00}
\definecolor{slout}{HTML}{0072B2}

\begin{figure*}[t]
  \centering
  \includegraphics[width=\textwidth]{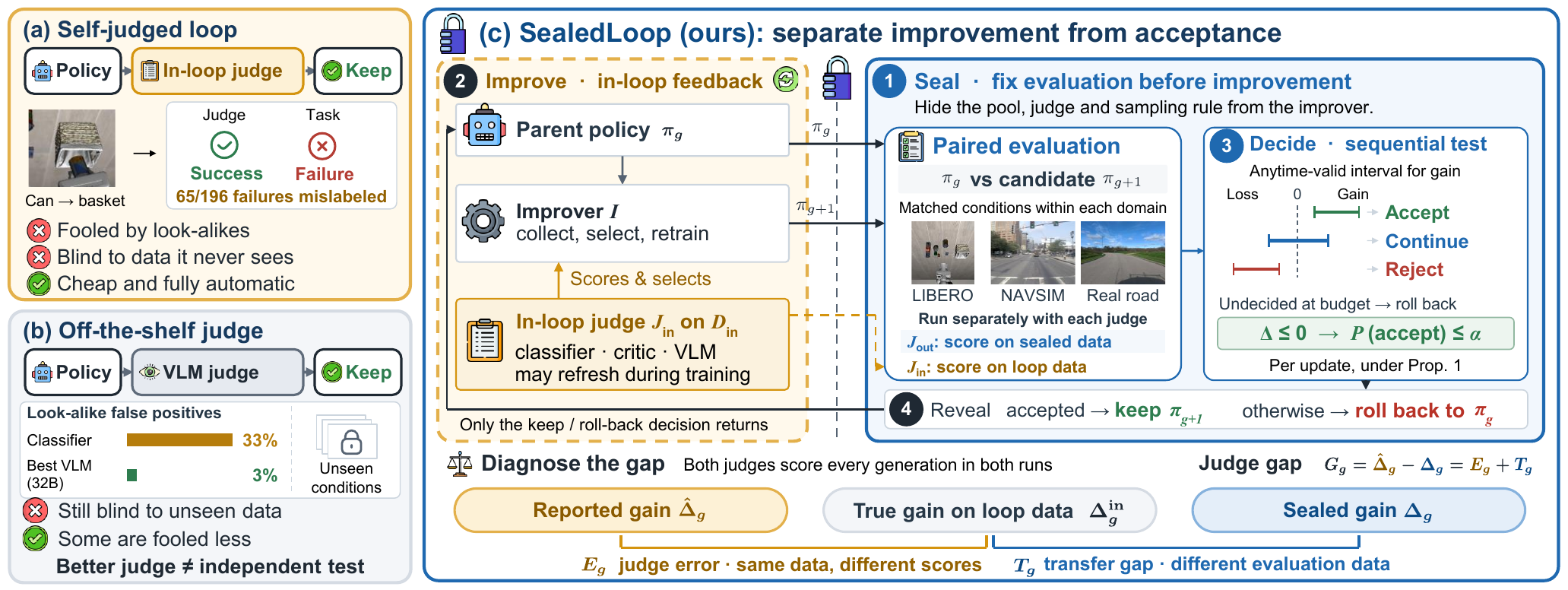}
  \caption{\textbf{Who decides what the loop keeps.} (a) The loop's own judge. (b) An off-the-shelf judge: fooled less, still blind to unseen data.
    (c) \method: the improver trains with $\Jin$ only, a sequential test on the sealed $\Jout$ decides, and both judges score each generation.}
  \label{fig:method}
\end{figure*}

\method adds a sealed judge and a sequential test on it to a self-improvement loop, and runs the loop once under each judge (\cref{fig:method}). We define the quantities the two judges measure, separate their disagreement into two sources, and give the test that decides each update.

\subsection{Problem setup}
\label{sec:setup_def}

A self-improvement loop holds a policy $\pi_g$ at generation $g$. An improver $\mathcal{I}$ produces a candidate $\pi_{g+1} = \mathcal{I}(\pi_g; \Jin)$ by collecting episodes and retraining with an \emph{in-loop judge} $\Jin$, such as a success classifier, a learned critic or a vision-language model. The improver may also update $\Jin$. The loop judges candidates on its own evaluation set $\Din$ and reports the gain
\begin{equation}
\hat{\Delta}_g = \Jin(\pi_{g+1}; \Din) - \Jin(\pi_g; \Din),
\end{equation}
where $\Jin(\pi; \Din)$ is the mean score the judge gives $\pi$ on $\Din$ and one version of the judge scores both policies. Let $s_c(\pi) \in [0, 1]$ be the true task score of $\pi$ on a condition $c$ (a success predicate, a rule-based driving score). A condition is a start state with a seed in manipulation and a log in driving, where $s_c$ is the mean PDMS over the frames of the log. Let $S(\pi; \mathcal{D})$ be the mean of $s_c(\pi)$ over the conditions of $\mathcal{D}$. A \emph{sealed judge} $\Jout$ applies $S$ on a pool $\Pool$ whose data, judge parameters and schedule are hidden from the improver. The \emph{sealed gain} and the \emph{judge gap} are
\begin{equation}
\Delta_g = S(\pi_{g+1}; \Pool) - S(\pi_g; \Pool), \qquad G_g = \hat{\Delta}_g - \Delta_g .
\end{equation}

\subsection{Decomposing the judge gap}
\label{sec:decomposition}

With $\Delta^{\mathrm{in}}_g = S(\pi_{g+1}; \Din) - S(\pi_g; \Din)$ the true gain on the loop's own data,
\begin{equation}
G_g = \underbrace{\hat{\Delta}_g - \Delta^{\mathrm{in}}_g}_{E_g\text{: judge error}} \; + \; \underbrace{\Delta^{\mathrm{in}}_g - \Delta_g}_{T_g\text{: transfer gap}} .
\label{eq:decomposition}
\end{equation}
We obtain $\Delta^{\mathrm{in}}_g$ by applying the true score $S$ to $\Din$; the loop itself does not use it. A better judge reduces $E_g$, but no judge that sees only $\Din$ can detect $T_g$, the part of a true gain that does not carry over to deployment conditions, such as an unseen city or a real vehicle. $T_g$ vanishes up to noise when $\Din$ and $\Pool$ share a distribution, as on manipulation.

\subsection{Certified acceptance}
\label{sec:gate}

The test draws conditions $c_1, c_2, \dots$ from $\Pool$, runs both policies on each, and reads the paired differences $d_i = s_{c_i}(\pi_{g+1}) - s_{c_i}(\pi_g) \in [-1, 1]$, whose mean is $\Delta_g$. A betting confidence sequence~\cite{waudbysmith2024betting} gives intervals $C_t = [L_t, U_t]$ from $d_1, \dots, d_t$ with
\begin{equation}
\Pr\left(\forall t \ge 1: \ \Delta_g \in C_t\right) \ge 1 - \alpha ,
\end{equation}
so the test may read $C_t$ after every pair~\cite{ramdas2023gametheoretic}. It \emph{accepts} if $L_t > 0$, \emph{rejects} if $U_t < 0$, and otherwise continues up to a budget $N$; an undecided candidate is rolled back. We call a gain \emph{certified} when the test accepts, and a loss certified when it rejects.

\begin{proposition}
\label{prop:gate}
Assume that each $d_i$ has mean $\Delta_g$ given $d_1, \dots, d_{i-1}$. Then, for any improver and any stopping time, the probability that the test accepts a candidate with $\Delta_g \le 0$ or rejects one with $\Delta_g \ge 0$ is at most $\alpha$.
\end{proposition}
\noindent\emph{Proof.} Acceptance at $t$ implies $L_t > 0 \ge \Delta_g$ and rejection implies $U_t < 0 \le \Delta_g$. In both cases $\Delta_g \notin C_t$, which has probability at most $\alpha$ over all $t$ jointly.

\noindent The assumption holds when the conditions are drawn independently of each other and of both policies. Sealing secures the second part: the conditions of a generation are fixed before the improver runs, so the improver cannot fit the candidate to them. Our draws are stratified without replacement. Under this sampling, the test accepts a candidate without a real gain in at most \SLPostSimInterWorst{} of simulated generations and none that loses 0.05 or 0.10 (\supp{Method details}). Over $K$ generations it makes at most $\alpha K$ wrong decisions in expectation under the same assumption.

\subsection{\method}
\label{sec:protocol}

\method combines the sealed judge and the test into four steps per generation (\cref{fig:method}):
\begin{enumerate}[leftmargin=*,itemsep=0pt,topsep=2pt]
\item \textbf{Seal:} fix the evaluation pool, the judge and the rule that draws this generation's conditions.
\item \textbf{Improve:} produce $\pi_{g+1}$ with $\Jin$ only; nothing from $\Jout$ reaches the improver.
\item \textbf{Decide:} evaluate $\pi_g$ and $\pi_{g+1}$ on paired conditions until the test accepts, rejects or reaches its budget.
\item \textbf{Reveal:} keep $\pi_{g+1}$ if it was accepted and roll back otherwise; record the readings of both judges.
\end{enumerate}
We run each loop in variants that differ in the judge that feeds the test in step~3. In the \emph{in-loop} variant the loop's own judge decides, and in the \emph{sealed} variant the sealed judge does. Both variants are scored by $\Jout$, so every generation yields $\hat{\Delta}_g$, $\Delta_g$ and both decisions, where a loop run once under its own judge yields only $\hat{\Delta}_g$. We count two kinds of wrong in-loop decisions: candidates kept with a negative sealed gain (\emph{worse kept}), and candidates rolled back although the sealed judge certifies their gain (\emph{gains lost}).

\section{Experimental Setup}
\label{sec:experiments}

\begin{table*}[t]
\centering\footnotesize
\setlength{\tabcolsep}{3.5pt}
\renewcommand{\arraystretch}{1.05}
\begin{tabular}{@{}>{\columncolor{white}[0pt][\tabcolsep]}lll>{\columncolor{white}[\tabcolsep][0pt]}l@{}}
\toprule
 & Manipulation (simulation) & Driving (simulation) & Real-world vehicle \\
\midrule
Benchmark & LIBERO-Object~\cite{libero2023}, \SLTasksSOne{} tasks & NAVSIM~\cite{navsim2024}, unseen city & full-scale van, three routes on a real road \\
Start policy & OpenVLA-OFT~\cite{openvlaoft2025}, few-shot adapted & DiffusionDrive~\cite{diffusiondrive2025}, reduced data & ReCogDrive-2B, imitation version (IL)~\cite{ax2506_08052} \\
Improver & rejection-sampling fine-tuning & best-of-\SLProposalsSThree{} self-distillation & published reinforcement-learning step (RL) \\
\rowcolor{black!8}In-loop judge & success classifier, refreshed on own labels & PDMS critic, refreshed on true or own labels & simulator score: NAVSIM \\
\rowcolor{black!8}Sealed judge & success predicate, unseen start conditions & PDMS in Boston & route completed before the first intervention \\
Runs & 3 tasks $\times$ 3 seeds $\times$ 3 variants & 3 seeds $\times$ 3 variants, 3 control runs & one step; blind paired drives of IL and RL \\
Per generation & in-loop \SLEvalPerGenInLoopSOne{}, sealed up to \SLEvalPerGen{} pairs & in-loop \SLInLoopFramesSThree{}, sealed \SLPoolFramesSThree{} frames & \SLVehLeftPairs{}, \SLVehStraightPairs{} and \SLVehRightPairs{} pairs on the three routes \\
\bottomrule
\end{tabular}
\caption{\textbf{Testbeds.} The variants of a testbed differ in which judge decides; on manipulation the two tests also differ in budget.}
\label{tab:testbeds}
\end{table*}

\rhead{Manipulation.}
The released OpenVLA-OFT checkpoints already solve LIBERO-Object (\cref{tab:testbeds}), so we adapt a LIBERO-Spatial checkpoint with a few demonstrations (\supp{Manipulation}); the three start policies score \SLStartsInBand{} on the sealed pool. Each generation collects episodes on \SLCollectPerGenSOne{} fresh start conditions per task and fine-tunes with LoRA on those the classifier labels successful. The classifier is refreshed on these labels, as in loops that learn it from their own episodes~\cite{ax2410_21845,ax2412_09858}. The sealed pool crosses the suite's 50 start states with environment seeds; each generation draws a disjoint set, stratified by start state.

\rhead{Driving.}
The start policy is DiffusionDrive trained from scratch on \SLImproverFramesSThree{} frames from Las Vegas, Pittsburgh and Singapore. It was chosen by its PDMS on \SLSubsetSelFrames{} Boston frames of the sealed pool; replaying every sealed test on the other \SLSubsetFrames{} frames gives the same decision for \SLSubsetSame{} of \SLSubsetCandidates{} candidates and accepts none (\supp{Driving}). Control runs start from the released checkpoint. The critic predicts PDMS, NAVSIM's rule-based driving score in $[0, 1]$, from the frozen encoder's features, as learned critics do in published driving loops~\cite{ax2508_09158,ax2604_27366}. It is refreshed every generation on true labels of fresh proposals or on its own pairwise preferences. Each generation, the critic picks one of \SLProposalsSThree{} proposals per training-city frame as the target, hard frames are replayed~\cite{ax2606_30537,ax2508_09158}, and the policy is retrained. The sealed judge reads PDMS on a frozen pool of \SLPoolFramesSThree{} Boston frames, a city no generation trains on.

\rhead{Real-world field experiment.}
The third testbed is a full-scale drive-by-wire electric van on a real road, with an RTK-corrected inertial navigation system and an on-board GPU (\cref{fig:vehicle}a). Both policy versions run through the same software, plan from the same front camera once per \SLVehPlanPeriod{}\,s, and are executed by the same trajectory tracker at a speed limit of 4\,m/s. Each camera frame is reprojected to the geometry of the NAVSIM front camera before it reaches the policy (\supp{Real-world field experiment}). The main route is \SLVehLeftLength{}\,m long, with a left curve of about 55$^\circ$; a \SLVehStraightLength{}\,m straight route and a \SLVehRightLength{}\,m route with a right curve complete the study. A pair is one drive of each version, back to back in a random order; the safety driver and the operator see only blind labels. All scored drives took place in November 2026. A drive's score is the fraction of the route completed before the first intervention (a brake, steering or throttle override, a timeout or a stall), read from the drive-by-wire records.

\rhead{Implementation details.}
All tests use $\alpha = \SLAlpha$, at least \SLMinN{} pairs and a threshold of zero. Runs last five generations, and \SLExtendedChainsSOne{} classifier-decided manipulation runs continue to eight. On driving the test reads the mean difference of each log. Improver processes run with the sealed judge's files masked.

\section{Experimental Results}
\label{sec:results}

\begin{figure*}[t]
\centering
\includegraphics[width=\textwidth]{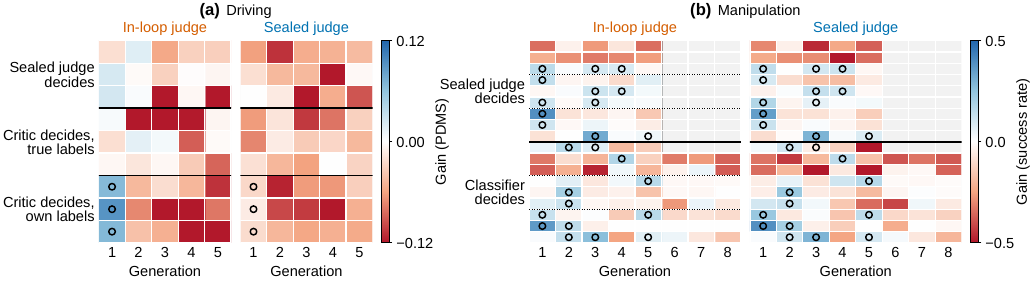}
\caption{\textbf{Every generation as seen by the two judges.} Each cell is a candidate's gain, as reported by the in-loop judge (left grid of each pair) and as measured by the sealed judge (right grid). Rows are runs; rings mark kept candidates.}
\label{fig:generations}
\end{figure*}

\subsection{Reported gains do not hold under a sealed judge}
\label{sec:r_diverge}

On driving, the two judges diverge at once (\cref{fig:teaser,fig:generations,tab:decisions}). At the first generation the judge gap is \SLJudgeGapGenOneInLoopRTwoSThree{} \SLJudgeGapGenOneCIInLoopRTwoSThree{} for the critic refreshed on its own labels and \SLJudgeGapGenOneInLoopROneSThree{} \SLJudgeGapGenOneCIInLoopROneSThree{} for the one refreshed on true labels. The former kept one candidate per seed, each scoring lower in Boston than the policy it replaced. The sealed judge rolled back all \SLSealedRollbacksSThree{} driving candidates (mean sealed gain at the first generation \SLSealedGainFirstMeanSThree{}). In this reduced-data setting the improver produced no gain that reached the unseen city, and the sealed judge shows it: with its own critic deciding, the same loop kept a worse policy in every seed.

\begin{table}[t]
\centering\small
\setlength{\tabcolsep}{3pt}
\begin{tabular}{@{}>{\columncolor{white}[0pt][\tabcolsep]}llccc>{\columncolor{white}[\tabcolsep][0pt]}c@{}}
\toprule
Testbed & Judge & Runs & Kept & Worse kept & Gains lost \\
\midrule
\ours Manipulation & sealed  & \SLSealedChainsSOne{} & \SLSealedAcceptedGensSOne{} & -- & -- \\
       & classifier & \SLInLoopChainsSOne{} & \SLInLoopAcceptedGensSOne{} & \zerogood{\SLInLoopAcceptedWorseSOne} & \zerogood{\SLFalseRollbacksSOne} \\
       & true success & \SLPredfixChains{} & \SLPredfixKept{} & \zerogood{\SLPredfixWorseKept} & \zerogood{\SLPredfixGainsLost} \\
\midrule
\ours Driving & sealed  & \SLSealedChainsSThree{} & \SLSealedAcceptedGensSThree{} & -- & -- \\
        & critic & \SLInLoopChainsSThree{} & \SLInLoopAcceptedGensSThree{} & \zerogood{\SLInLoopAcceptedWorseSThree} & \zerogood{\SLFalseRollbacksSThree} \\
\bottomrule
\end{tabular}
\caption{\textbf{Acceptance decisions} in generations 1 to 5 of each run.}
\label{tab:decisions}
\end{table}

On manipulation the improver works, and the in-loop judge still makes both kinds of wrong decision. The judges diverge later than on driving: in the median run that reaches this threshold, the gap first exceeds \SLGapThresholdSOne{} at generation \SLDivergenceGenNumSOne{}. The sealed judge kept \SLSealedAcceptedGensSOne{} generations (mean certified gain \SLSealedGainLastSOne{}); the classifier kept one candidate certified as worse and rolled back \SLFalseRollbacksSOne{} certified gains. Both variants end with better policies: re-run on new seeds, the final policies gain \SLPostAuditSealedFinalAMean{} (sealed) and \SLPostAuditInLoopFinalAMean{} (classifier) over their start policies on average, and \SLPostAuditSealedFinalBMean{} and \SLPostAuditInLoopFinalBMean{} on new start states, where one kept candidate of each variant loses (\supp{Manipulation}). Over the \SLPostBudgetInLoopN{} classifier-decided generations, replacing the classifier by the success predicate on the same pairs changes \SLPostBudgetInLoopJudge{} decisions and replacing \SLEvalPerGenInLoopSOne{} pairs by the sealed budget changes \SLPostBudgetInLoopBudget{}, so the judge and the amount of evaluation each change decisions. A third variant runs this comparison as a loop, in which the true success of the same \SLEvalPerGenInLoopSOne{} pairs decides. Its decisions match the sealed test in \SLPredfixSameAsSealed{} of \SLPredfixGens{} generations (\cref{tab:decisions}), so on manipulation the wrong decisions come from the classifier's labels.

\begin{figure*}[t]
\centering
\includegraphics[width=\textwidth]{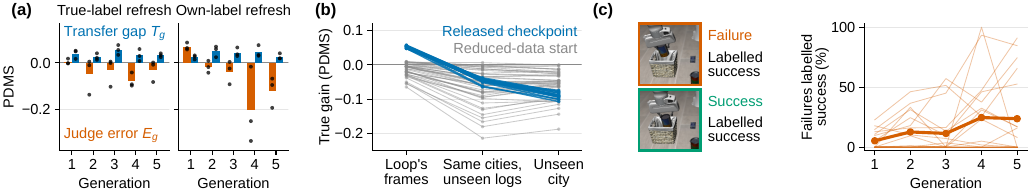}
\caption{\textbf{Two sources of the judge gap.} (a) Driving: judge error $E_g$ and transfer gap $T_g$ per generation under the two critic refresh rules (bars: three-seed mean; dots: seeds). (b) True gain of each driving candidate on three sets of frames. (c) Manipulation: a failure and a success, both labeled success by the classifier, and the share of each candidate's failures so labeled (thin lines: runs; thick line: mean).}
\label{fig:sources}
\end{figure*}

\subsection{The gap has two sources}
\label{sec:r_sources}

\rhead{Judge error.}
The in-loop judge is reliable on the data it was fitted to and loses that reliability as the loop moves the policy and the data away from it. Before the first generation the critic's pick beats the policy's own choice by \SLCriticSelTrainSThree{} PDMS on the loop's frames; in Boston its false-positive rate rises from \SLCriticFPRTrainSThree{} to \SLCriticFPRHeldSThree{} and its pick is worse (\SLCriticSelHeldSThree{}; \supp{Driving}). Refreshed on its own labels, the critic's error is \SLJudgeErrorGenOneRTwoSThree{} \SLJudgeErrorGenOneRTwoCISThree{} at the first generation, where it keeps worse policies (\cref{fig:sources}a); later the refreshed critics turn pessimistic. On manipulation, $\Din$ and $\Pool$ share one distribution, so the gap is judge error. The classifier's false-positive rate rises with the policy's own failures (\cref{fig:sources}c; across runs, \SLJudgeFPRStartSOne{} at the first generation and \SLJudgeFPREndSOne{} at the last), in two patterns. The new policy can fail with the object held at the edge of the basket, which looks like a success; the classifier accepted \SLFooledAcceptedFailuresSOne{} failures in one such generation. In \SLBlindGenerationsSOne{} generations of one run, a refresh without failures blinded the classifier to the next policy's failures (\supp{Manipulation}). Elsewhere the refresh changes few decisions: with the classifier frozen before the loop, \SLAddFlips{} of \SLAddFlipsOf{} differ and no run keeps a different policy (\supp{Additional analyses}).

\rhead{Transfer gap.}
Even a judge without error would keep worse policies, because gains on the loop's own data do not carry over to data the loop has not seen. In every driving generation and seed, the true gain on the loop's frames exceeds the gain in Boston (\SLTransferRangeMainSThree{}). Runs from the released checkpoint change their score by \SLControlVinRangeSThree{} on the loop's evaluation frames and by \SLControlPOneRangeSThree{} in Boston (\cref{fig:sources}b). The loss does not need a new city: on unseen logs of the training cities the same candidates change by \SLPostTransferCtlPThreeRange{}, \SLPostTransferCtlShare{} of the drop from the loop's frames to Boston, and \SLPostTransferMainNeg{} of the \SLPostTransferMainN{} candidates of the main runs lose there too. Nor does it need judge error: fed with the \emph{true} scores on the loop's evaluation frames, the test accepts \SLVinReplayBlockAcceptCtlSThree{} control generations, but none of the \SLPThreeReplayN{} candidates on the unseen logs. Nor do better targets remove it: trained on the proposal with the highest true score on each frame (\SLAddOracleTargetPDMS{} on average against \SLAddOraclePolicyPDMS{} for the policy's own choice), the first-generation candidate still changes its Boston score by \SLAddOracleGainLow{}, \SLAddOracleGainMid{} and \SLAddOracleGainHigh{} at three learning rates (\supp{Driving}). The transfer gap comes from evaluating on the loop's data; only a sealed pool exposes it.

\subsection{Off-the-shelf judges do not replace a sealed one}
\label{sec:r_stronger}

\begin{figure}[t]
\centering
\includegraphics[width=\linewidth]{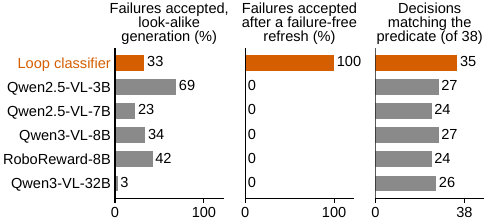}
\caption{\textbf{Off-the-shelf judges on manipulation.} Share of failures accepted on a look-alike generation (\SLJudgesHardN{} failures) and after a failure-free refresh (\SLJudgesBlindN{}), and generations where the test fed by each judge matches the success predicate.}
\label{fig:judges}
\end{figure}

A stronger judge is the obvious remedy for judge error, yet none of the off-the-shelf judges we tried replaces the sealed one. We re-scored the recorded manipulation episodes with four off-the-shelf vision-language models and a robot reward model~\cite{ax2601_00675} (\cref{fig:judges}). None is blinded by the failure-free refresh, which affects only the judge trained in the loop. On the generation whose failures look like successes, only \SLJudgesHardFewer{} of the \SLJudgesOthers{} accept fewer failures than the classifier (\SLJudgesHardJThree{} and \SLJudgesHardJSix{} against \SLJudgesHardJZero{}), and \SLJudgesHardMore{} accept more. No judge reaches the success predicate's decision on the same pairs as often as the loop's classifier (\SLLadderAgreeJZero{} of \SLLadderPrimaryGens{} generations against at most \SLLadderBestVlmAgree{}). The largest judge, Qwen3-VL-32B, accepts the fewest failures but also rejects many true successes (true-positive rate \SLLadderTPRJSix{}) and agrees on \SLLadderAgreeJSix{} generations. We also ran the loop with a vision-language judge deciding, in three runs. It kept \SLVLMCertificatesSOne{} of \SLVLMGenerationsSOne{} candidates and no worse policy, but rolled back \SLVLMFalseRollbacksSOne{} certified gains, and its decisions matched those of the success predicate on \SLVLMSameAsPredicateSOne{} generations against \SLCZeroSameAsPredicateVLMArmSOne{} for a frozen classifier. A larger judge thus trades accepted failures for rejected successes. \supp{Manipulation} gives the inputs and thresholds of every judge.

\subsection{Certified acceptance keeps gains, not losses}
\label{sec:r_gate}

\begin{table}[t]
\centering\small
\setlength{\tabcolsep}{1.8pt}
\begin{tabular}{@{}>{\columncolor{white}[0pt][\tabcolsep]}llccc>{\columncolor{white}[\tabcolsep][0pt]}r@{}}
\toprule
Judge & Rule & Kept & Worse kept & Regret & Frames \\
\midrule
\ours Sealed & sequential test (ours) & 0 & \good 0 & \good 0.000 & 354{,}754 \\
 & first 17 logs, read once & 0 & \good 0 & \good 0.000 & 41{,}160 \\
 & whole pool & 0 & \good 0 & \good 0.000 & 750{,}825 \\
In-loop & sequential test & 3 & \bad 3 & \bad 0.061 & n/a \\
 & peak of its report & 8 & \bad 8 & \bad 0.267 & n/a \\
\bottomrule
\end{tabular}

\caption{\textbf{Acceptance rules replayed on the driving runs.} Worse kept, regret: number and total sealed loss of kept candidates with negative sealed gain. In-loop test: six critic-decided runs.}
\label{tab:rules}
\end{table}

\begin{figure*}[t]
\centering
\includegraphics[width=\linewidth]{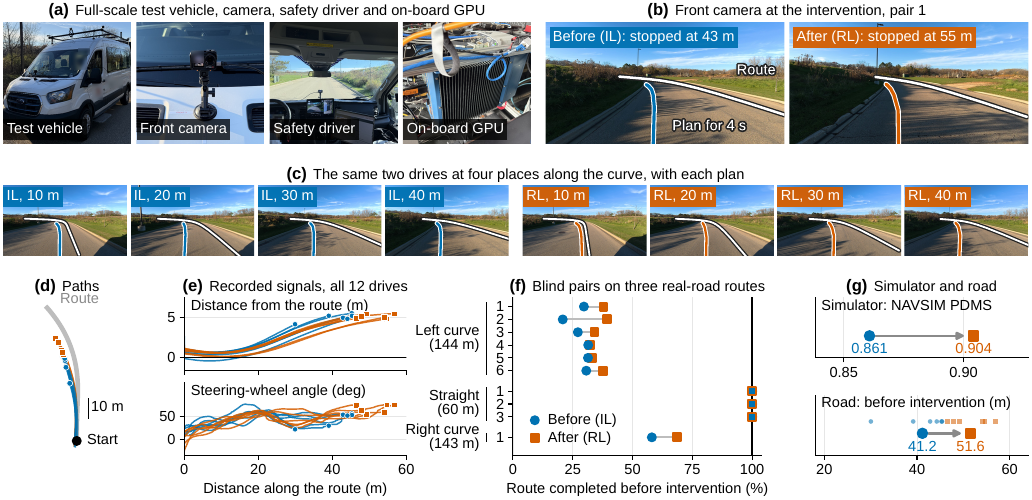}
\caption{\textbf{Field experiment on a full-scale vehicle.} (a) Test vehicle. (b, c) Front camera of the first pair at the intervention and before it, with planned trajectories. (d, e) Paths, distance from the route and steering angle on the left curve. (f) Route completed before the first intervention per blind pair. (g) The same step in the simulator and on the left curve (large markers: means; small: pairs).}
\label{fig:vehicle}
\end{figure*}

\begin{figure*}[t]
\centering
\includegraphics[width=\linewidth]{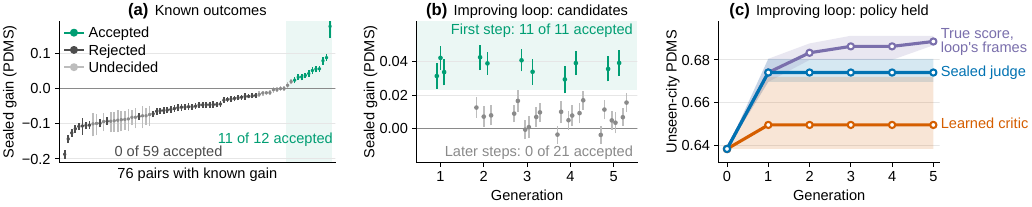}
\caption{\textbf{What the sequential test accepts.} (a) \SLPostCtlPairs{} policy pairs with known sealed gain (95\% intervals), colored by the test's decision. (b) Candidates of a loop built to improve. (c) Unseen-city PDMS of the policy it holds under three judges (line: seed mean; band: range).}
\label{fig:gate}
\end{figure*}

\rhead{Known gains.}
The sealed test accepts gains where they exist. It kept \SLSealedAcceptedGensSOne{} manipulation generations and no driving one, since no driving candidate gained in the unseen city. On \SLPostCtlPairs{} pairs of driving policies whose sealed gain is known from the whole pool (\cref{fig:gate}a), it accepts \SLPostCtlSeqGainsAcc{} of the \SLPostCtlGains{} real gains and none of the \SLPostCtlLosses{} losses or the \SLPostCtlNearZero{} pairs near zero. Averaged over \SLFairOrders{} log orders, a bootstrap over 50 logs fixed in advance accepts \SLFairBootFiftyGains{} of the \SLFairGains{} per-log gains with \SLFairBootFiftyShare{} of the frames; the sequential test accepts \SLFairSeqGains{} with \SLFairSeqShare{}. What the test adds is an error rate that holds wherever it stops: on pools built to have no gain, the same bootstrap read after every \SLFairPeekEvery{} logs accepts \SLFairPeekNullFA{}, against a nominal 2.5\%, and the sequential test accepts none (\supp{Driving}).

\rhead{A loop built to improve.}
When a loop does improve, the sealed test certifies its gains, and the judge that loses them is the loop's own. A start policy trained on 5{,}000 frames imitates the logged expert on 20{,}000 new frames per generation. Of the \SLWorkCands{} candidates, \SLWorkFirst{} are a first step from the start policy and gain \SLWorkAccGainLo{} to \SLWorkAccGainHi{} PDMS in the unseen city. The sealed test accepted all \SLWorkFirst{} (\cref{fig:gate}b); the learned critic accepted \SLWorkFirstCriticAcc{}. The other \SLWorkLater{} build on an accepted policy and gain \SLWorkUndGainLo{} to \SLWorkUndGainHi{}; the sealed test left all of them undecided. The held policy reaches \SLWorkEndPDMS{} from \SLWorkStartPDMS{} under the sealed judge, \SLWorkCriticEndPDMS{} under the critic, and \SLWorkVinEndPDMS{} under a paired test on the true scores of the loop's own frames (\cref{fig:gate}c). Judge error thus costs in both directions. The paired test on true scores keeps more here because these gains transfer; on the self-training runs above the same evidence kept policies that lose.

\rhead{Acceptance rules.}
Which judge feeds the rule matters more than the rule itself (\cref{tab:rules}). Rules fed by the sealed judge keep no worse policy. Fed by the in-loop judge, the same test and the rollback-to-peak rule proposed for language models~\cite{ax2606_21090} each keep some. The sequential test reaches the decisions of reading the whole pool every generation with \SLSeqShareMainSThree{} of its frames.

\subsection{Field experiment on a full-scale vehicle}
\label{sec:r_vehicle}

\rhead{Tested step.}
Finally, in a real-world field experiment, we test whether the simulator gain of one published fine-tuning step shows up on a real vehicle. ReCogDrive~\cite{ax2506_08052} fine-tunes an imitation-learned driving policy (IL) by reinforcement learning (RL) whose reward is the NAVSIM score, and it reports its gain with the same simulator and metric, on the separate navtest split. On our pipeline the two released versions score \SLVehicleNavsimILSFour{} and \SLVehicleNavsimRLSFour{} on navtest, a paired gain of \SLVehicleNavsimPairedSFour{} over \SLVehicleNavsimTokensSFour{} scenes. The sealed judge is the full-scale vehicle of \cref{sec:experiments}: both versions drove in blind pairs, and each drive is scored by the fraction of the route completed before the first intervention.

\rhead{On the road.}
No drive completed the main route (\SLVehLeftCompleted{}). Both versions are deployed zero-shot, without fine-tuning on data from our vehicle, so the interventions reflect a deployment gap that both share. The safety driver stopped every drive in the first half of the curve, between \SLVehLeftFailMin{} and \SLVehLeftFailMax{}\,m (\cref{fig:vehicle}d). In both versions the planned trajectory leaves the route toward the inner curb of the curve (\cref{fig:vehicle}b, c; paths and steering in \cref{fig:vehicle}d, e; \supp{Real-world field experiment}). RL went further into the curve in \SLVehLeftRLFurther{} pairs (exact sign test, one-sided $p = \SLVehLeftSignP{}$), by \SLVehLeftMeanDiffM{}\,m on average (\SLVehLeftMeanDiff{} of the route; \cref{fig:vehicle}f). The sequential test, which bounds the size of the gain, is undecided after \SLVehLeftPairs{} pairs (confidence sequence \SLVehLeftCS{}; \supp{Real-world field experiment}). On the straight route both versions completed every drive (\SLVehStraightCompleted{}), and on the right curve RL went further in the one pair driven. The direction of the simulator gain held on the road in all \SLVehLeftPairs{} pairs: the step that raises the score from \SLVehicleNavsimILSFour{} to \SLVehicleNavsimRLSFour{} moves the intervention from \SLVehLeftMeanIL{} to \SLVehLeftMeanRL{}\,m along the \SLVehLeftLength{}\,m route (\cref{fig:vehicle}g).

\section{Conclusion}
\label{sec:conclusion}

Self-improving embodied policies, in driving and in manipulation, can report improvement that a judge outside the loop does not find, because the in-loop judge errs on the loop's own data and real gains on that data fail to transfer to unseen data. Stronger off-the-shelf judges remove neither. \method instead seals the deciding judge from the improver. The sealed test accepted \SLPostCtlSeqGainsAcc{} of \SLPostCtlGains{} real gains and none of \SLPostCtlLosses{} losses, and certified all \SLWorkFirst{} first-step gains of a loop built to improve, of which the loop's own critic accepted \SLWorkFirstCriticAcc{}. In our real-world field experiment, the direction of a gain measured in simulation held on the road: the reinforced version drove \SLVehLeftMeanDiffM{}\,m further into a curve that neither version completed.

A separate test set is not enough, since our in-loop variants had the sealed reading and still kept worse policies. The sealed judge must therefore make the decision itself, and a sequential test lets it do so at \SLSeqShareMainSThree{} of the evaluation cost. So far, our loops run in simulation, and the field experiment tests one published improvement step on one vehicle. Future work will run the loop on hardware, with a simulator as the in-loop judge and the physical platform as the out-of-loop evaluator.

\section*{Acknowledgment}
This research was supported by the U.S. Department of Transportation (No.~69A3552348305) and by grants from NVIDIA utilizing NVIDIA RTX PRO 6000 Blackwell Max-Q Workstation Edition GPUs.

\section*{LLM Usage Statement}
We used Claude Code (Anthropic), a coding assistant built on large language models, as a coding and writing assistant in this work. It helped write and review experiment scripts, prepare figures and tables from the recorded results, and edit the text for clarity. All experimental designs, hypotheses and conclusions were decided by the authors, every reported number is produced by scripts from the sealed records, and the authors take full responsibility for the content of the paper.

{
    \small
    \bibliographystyle{ieeenat_fullname}
    \bibliography{main}
}
\clearpage
\appendix
\section{Method details}
\label{sec:supp_method}

\Cref{alg:supp_loop} lists one generation of the sealed variant, with $n_0 = \SLMinN$. In the in-loop variant the differences of line~5 come from $\Jin$ on $\Din$, and line~12 is unchanged.

\subsection{The sequential test}
\rhead{The confidence sequence.}
The test of \cref{sec:gate} reads paired differences $d_1, d_2, \dots \in [-1, 1]$ and tests their mean $\Delta$ (we drop the generation index). Let $u_i = (d_i + 1)/2 \in [0, 1]$, and for a candidate mean $m \in (-1, 1)$ let $\tilde m = (m + 1)/2$. The bets are computed from earlier pairs only:
\begin{equation}
\begin{aligned}
\hat\mu_{i-1} &= \frac{\tfrac12 + \sum_{j<i} u_j}{i}, \\
\hat\sigma^2_{i-1} &= \frac{\tfrac14 + \sum_{j<i} (u_j - \hat\mu_{j-1})^2}{i}, \\
\lambda_i &= \sqrt{\frac{2 \log (2/\alpha)}{\hat\sigma^2_{i-1}\, i \log (1 + i)}} .
\end{aligned}
\end{equation}
Each bet is truncated with $c = 0.75$,
\begin{equation}
\lambda^+_i(m) = \min\{\lambda_i, c/\tilde m\}, \quad \lambda^-_i(m) = \min\{\lambda_i, c/(1 - \tilde m)\},
\end{equation}
and the capital of a bettor against the candidate $m$ is
\begin{equation}
\begin{aligned}
K_t(m) = {} & \frac12 \prod_{i=1}^{t} \bigl(1 + \lambda^+_i(m)(u_i - \tilde m)\bigr) \\
 & + \frac12 \prod_{i=1}^{t} \bigl(1 - \lambda^-_i(m)(u_i - \tilde m)\bigr).
\end{aligned}
\label{eq:supp_capital}
\end{equation}
The confidence sequence keeps the candidates against which the capital has stayed small~\cite{waudbysmith2024betting}:
\begin{equation}
C_t = \Bigl\{ m : \max_{r \le t} K_r(m) < 1/\alpha \Bigr\},
\label{eq:supp_cs}
\end{equation}
with $L_t = \inf C_t$ and $U_t = \sup C_t$.

\begin{algorithm}[t]
\caption{One generation of \method}
\label{alg:supp_loop}
\footnotesize
\begin{algorithmic}[1]
\Require policy $\pi_g$, improver $\mathcal{I}$, in-loop judge $\Jin$ with data $\Din$, sealed pool $\Pool$ with true score $s$, level $\alpha$, budget $N$, first reading $n_0$
\State \textbf{Seal:} fix $\Pool$, $s$ and the rule that draws $c_1, c_2, \dots$ from $\Pool$
\State \textbf{Improve:} $\pi_{g+1} \gets \mathcal{I}(\pi_g; \Jin)$ \Comment{no access to $\Pool$, $s$ or the draws}
\State \textbf{Decide:} decision $\gets$ undecided
\For{$t = 1, \dots, N$}
  \State run $\pi_g$ and $\pi_{g+1}$ on $c_t$; \ $d_t \gets s_{c_t}(\pi_{g+1}) - s_{c_t}(\pi_g)$
  \State compute $[L_t, U_t]$ from $d_1, \dots, d_t$ \Comment{\cref{eq:supp_cs}}
  \If{$t \ge n_0$ and $L_t > 0$} decision $\gets$ accept; \textbf{break}
  \ElsIf{$t \ge n_0$ and $U_t < 0$} decision $\gets$ reject; \textbf{break}
  \EndIf
\EndFor
\State \textbf{Reveal:} keep $\pi_{g+1}$ if decision is accept, otherwise roll back to $\pi_g$; the improver receives only this outcome
\State record $\hat\Delta_g$ from $\Jin$ on $\Din$ and $\Delta_g$ from $s$ on $\Pool$
\end{algorithmic}
\end{algorithm}

\begin{supplemma}[Coverage]
\label{lem:supp_cover}
If $\mathbb{E}[d_i \mid d_1, \dots, d_{i-1}] = \Delta$ for every $i$, then $\Pr(\exists\, t \ge 1: \Delta \notin C_t) \le \alpha$.
\end{supplemma}
\begin{proof}
Set $m = \Delta$. Since $u_i - \tilde m \ge -\tilde m$ and $\lambda^+_i \le c/\tilde m$, every factor of the first product in \cref{eq:supp_capital} is at least $1 - c > 0$, and likewise for the second. The bets depend on $d_1, \dots, d_{i-1}$ only and $\mathbb{E}[u_i - \tilde m \mid d_1, \dots, d_{i-1}] = 0$, so each factor has conditional mean one. $K_t(\Delta)$ is therefore a nonnegative martingale with $K_0 = 1$, and Ville's inequality~\cite{ramdas2023gametheoretic} gives $\Pr(\sup_t K_t(\Delta) \ge 1/\alpha) \le \alpha$.
\end{proof}
\noindent \Cref{prop:gate} follows: an acceptance at $t$ with $\Delta \le 0$ or a rejection with $\Delta \ge 0$ implies $\Delta \notin C_t$. The bound holds for any rule that chooses when to read $C_t$, including a first reading at pair \SLMinN{} and a stop at the budget.

\begin{supplemma}[Sealing gives the assumption]
\label{lem:supp_seal}
Let the conditions $c_1, c_2, \dots$ be drawn independently from a distribution $P$ on the pool, independently of $(\pi_g, \pi_{g+1})$, and let the two policies be run on each condition with fresh randomness. Then $\mathbb{E}[d_i \mid d_1, \dots, d_{i-1}] = \Delta_g$ with $\Delta_g = \mathbb{E}_{c \sim P}[s_c(\pi_{g+1}) - s_c(\pi_g)]$.
\end{supplemma}
\begin{proof}
Given the two policies, the pairs $d_1, d_2, \dots$ are independent and identically distributed with mean $\Delta_g$, so conditioning on earlier pairs does not change the mean of $d_i$.
\end{proof}
\noindent The independence from $\pi_{g+1}$ is what sealing provides. An improver that had seen the conditions could fit the candidate to them; the $d_i$ would then have a mean above the gain on new conditions, and the test would certify a gain that exists only on the conditions it reads.

\begin{suppcorollary}[Several generations]
\label{cor:supp_gens}
Let a loop run $K$ generations, each decided by the test at level $\alpha$, and let the assumption of \cref{lem:supp_cover} hold in every generation given all earlier generations. Let $W$ be the number of generations in which the test accepts a candidate with $\Delta_g \le 0$ or rejects one with $\Delta_g \ge 0$. Then $\mathbb{E}[W] \le \alpha K$ and $\Pr(W \ge 1) \le \alpha K$.
\end{suppcorollary}
\begin{proof}
Given the earlier generations, generation $g$ is decided wrongly with probability at most $\alpha$ by \cref{prop:gate}. Taking expectations and summing over $g$ gives $\mathbb{E}[W] \le \alpha K$, and $\Pr(W \ge 1) \le \mathbb{E}[W]$.
\end{proof}

\begin{supproposition}[An in-loop rule cannot see the transfer gap]
\label{prop:supp_transfer}
Let the pool contain no condition of $\Din$, and let an acceptance rule, possibly randomized, depend only on what the two policies do on $\Din$. Then for any such data there are two score functions on the pool, one with $\Delta_g > 0$ and one with $\Delta_g < 0$, under which the rule behaves identically. If it accepts with probability $p$, it accepts a loss with probability $p$ in one case and rolls back a gain with probability $1 - p$ in the other, so the two error probabilities sum to one.
\end{supproposition}
\begin{proof}
The rule's input does not involve $s_c$ for $c$ in the pool. Choose $s_c(\pi_{g+1}) - s_c(\pi_g) = \delta$ on every pool condition in the first case and $-\delta$ in the second, for some $\delta \in (0, 1]$. The input, and hence the distribution of the decision, is the same in both.
\end{proof}
\noindent In the decomposition of \cref{eq:decomposition}, the two cases have the same $\hat\Delta_g$ and $\Delta^{\mathrm{in}}_g$, hence the same judge error $E_g$, and differ only in the transfer gap $T_g$. No improvement of the in-loop judge separates them; a reading on the pool does.

\rhead{Size of a certified gain.}
An acceptance at a stopping time $\tau$ certifies the sign of $\Delta_g$. The mean of the pairs read up to $\tau$ overstates its size, because the test stops when the differences seen so far are large. The interval $[L_\tau, U_\tau]$ remains valid by \cref{lem:supp_cover}, and we report sealed gains from all pairs of the generation or from the whole pool.

\subsection{Parameters, sampling and sealing}
\rhead{Test parameters.}
Every test in the paper uses the betting confidence sequence~\cite{waudbysmith2024betting} at $\alpha = \SLAlpha$ with a threshold of zero. Bets are truncated as in that method, and the test reads its interval after every pair from pair \SLMinN{} on. A candidate undecided at the budget is not kept.

\rhead{Pairs and sampling.}
On manipulation, a pair is one start condition (one of the task's 50 start states with an environment seed) on which both policies are run. The sealed pool of a task holds 200 seeds per start state, 10{,}000 conditions in all. Each generation draws a new set that is disjoint from all earlier ones and stratified by start state, up to a budget of \SLEvalPerGen{} pairs. On driving, a pair is one log scored for both policies, and its paired difference is the mean over the frames of the log. The test reads the logs of the sealed pool once, in a fixed pseudo-random order. We also run the test on single frames; frames of one log are correlated, so the per-log result decides when the two disagree. In a bootstrap over logs, the per-log test detects a gain of about \SLBlockMDEPTwoSThree{} PDMS with probability 0.8 on the main pool of 280 logs.

\begin{table}[t]
\centering\footnotesize
\setlength{\tabcolsep}{4pt}
\begin{tabular}{@{}>{\columncolor{white}[0pt][\tabcolsep]}lcc>{\columncolor{white}[\tabcolsep][0pt]}c@{}}
\toprule
 & \multicolumn{2}{c}{False acceptances} & $\geq 1$ in five \\
\cmidrule(lr){2-3}
Sampling & Worst setting & Pooled & generations \\
\midrule
\rowcolor{black!8}\multicolumn{4}{@{}>{\columncolor{white}[0pt][0pt]}l@{}}{\emph{Manipulation}} \\
By start state (ours) & \SLPostSimInterWorst{} & \SLPostSimInterPooled{} & \SLPostSimInterRun{} \\
Uniform, no replacement & \SLPostSimUnifWorst{} & \SLPostSimUnifPooled{} & \SLPostSimUnifRun{} \\
With replacement & \SLPostSimIidWorst{} & \SLPostSimIidPooled{} & \SLPostSimIidRun{} \\
\midrule
\rowcolor{black!8}\multicolumn{4}{@{}>{\columncolor{white}[0pt][0pt]}l@{}}{\emph{Driving}} \\
Per-log result decides (ours) & & \SLPostSimDriveFA{} of \SLPostSimDriveN{} & \\
Per-frame test alone & & \SLPostSimDriveFrameFA{} of \SLPostSimDriveN{} & \\
\bottomrule
\end{tabular}
\caption{\textbf{False acceptances under the sampling used.} Share of simulated generations without a real gain that the test accepts. Last column: at least one acceptance in five such generations, at the worst setting's rate.}
\label{tab:supp_sampling}
\end{table}

\rhead{Error rate under the sampling used.}
\Cref{prop:gate} and \cref{cor:supp_gens} assume independent pairs (\cref{lem:supp_seal}). Our manipulation pairs are drawn without replacement from a finite pool, stratified by start state, and driving frames share logs, so we measure the error rate by simulation (\cref{tab:supp_sampling}). Manipulation pools are built from the measured success rate of every start state and the measured agreement between seeds (\SLPostSimCells{} settings, \SLPostSimReplicates{} simulated generations); on driving, the two scores of a pair of start policies are swapped in a random half of the logs. With our draw, the test accepts a candidate without a real gain in at most \SLPostSimInterWorst{} of generations in any setting, below $\alpha$; five such generations contain an acceptance with probability \SLPostSimInterRun{}, against a union bound of \SLPostSimUnionBound{}. Under all three draws, no candidate that loses 0.05 or 0.10 is accepted (\SLPostSimLossAcc{} of \SLPostSimLossRuns{}), and a gain of 0.03 is accepted in \SLPostSimSmallGainAcc{} of generations at \SLEvalPerGen{} pairs. On driving, the per-frame test alone accepts \SLPostSimDriveFrameFA{} of \SLPostSimDriveN{} pools without a gain and the per-log test none. Replayed with a confidence sequence for sampling without replacement~\cite{waudbysmith2020wor}, every recorded acceptance and rejection that can be re-read is unchanged (\SLWorReplayManipSame{} on manipulation and \SLWorReplayDriveSame{} on driving). Manipulation draws are disjoint across generations, so no earlier decision depends on the conditions a generation reads; on driving every generation reads the same frozen pool, from which the improver learns one decision per generation. The error rate of the test therefore stays below $\alpha$ under the sampling we use.

\rhead{Isolation of the sealed judge.}
Every improver process runs with the files of the sealed judge masked from its file system, and the evaluation conditions of the two judges are disjoint. On driving, all \SLIsolationStepsSThree{} improver steps passed a check of file access; on manipulation a check of every generation found no access to them (\SLIsolationPosthocSOne{}).

\section{Manipulation}
\label{sec:supp_manip}

\begin{figure*}[t]
\centering
\includegraphics[width=\textwidth]{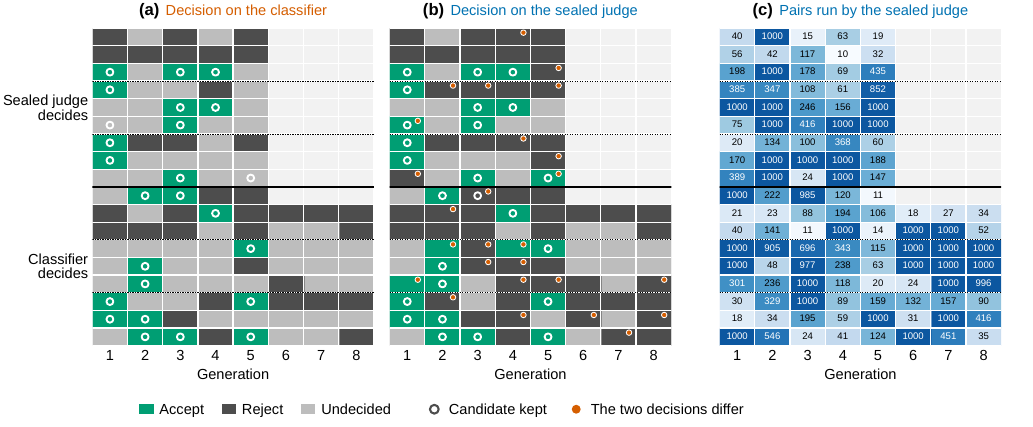}
\caption{\textbf{Decisions in every manipulation generation.} Rows are the eighteen runs in the order of \cref{fig:generations}, with dashed lines between tasks. (a) Decision of the test on the classifier's verdicts. (b) Decision on the sealed judge; an orange dot marks the \SLSuppManipDiffer{} generations in which the two differ. (c) Pairs the sealed judge ran before the test stopped.}
\label{fig:supp_manip_decisions}
\end{figure*}

\begin{figure}[t]
\centering
\includegraphics[width=\linewidth]{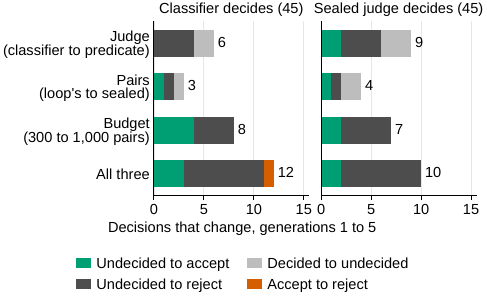}
\caption{\textbf{Judge, pairs and budget.} Decisions of the first five generations that change when the judge, the pairs, the budget or all three are replaced by those of the sealed test.}
\label{fig:supp_factors}
\end{figure}

\begin{figure}[t]
\centering
\includegraphics[width=\linewidth]{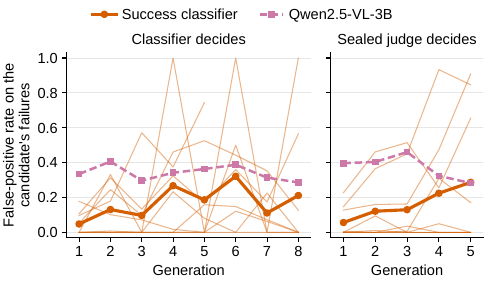}
\caption{\textbf{False-positive rate of the classifier by generation.} Share of the candidate's failed episodes that the classifier labels as successes (thin lines: runs; thick line: mean). Dashed: the same rate for Qwen2.5-VL-3B, which is not trained in the loop.}
\label{fig:supp_fpr}
\end{figure}

\subsection{Tasks, policies and judges}
The testbed uses three LIBERO-Object tasks~\cite{libero2023}: placing the alphabet soup, the cream cheese and the BBQ sauce in the basket (tasks 1 to 3). The source model is the OpenVLA-OFT checkpoint for LIBERO-Spatial~\cite{openvlaoft2025}. For each task we adapt it with LoRA on a few demonstrations and take the checkpoint whose success on calibration conditions lies between 0.50 and 0.80. On \SLEvalPerGen{} sealed conditions the three start policies succeed at rates of \SLStartsInBand{}. In every generation the policy runs \SLCollectPerGenSOne{} new start conditions per task, the in-loop judge selects the episodes it rates as successful, and the LoRA adapter is trained further on these episodes alone (rank 32, 2{,}000 steps, batch size 8, learning rate $10^{-4}$). The in-loop judge is a success classifier that reads the final camera frames of an episode and is retrained every generation on its own labels. Its test reads \SLEvalPerGenInLoopSOne{} pairs on conditions excluded from collection. The sealed judge is the simulator's success predicate on the sealed pool of \cref{sec:supp_method}. Three seeds per task and variant give eighteen runs of five generations. The classifier-decided runs that had kept no candidate certified as worse by the sealed judge continued to eight generations (\SLExtendedChainsSOne{} runs, \SLExtensionGenerationsInLoopSOne{} generations).

\begin{figure*}[t]
\centering
\includegraphics[width=\textwidth]{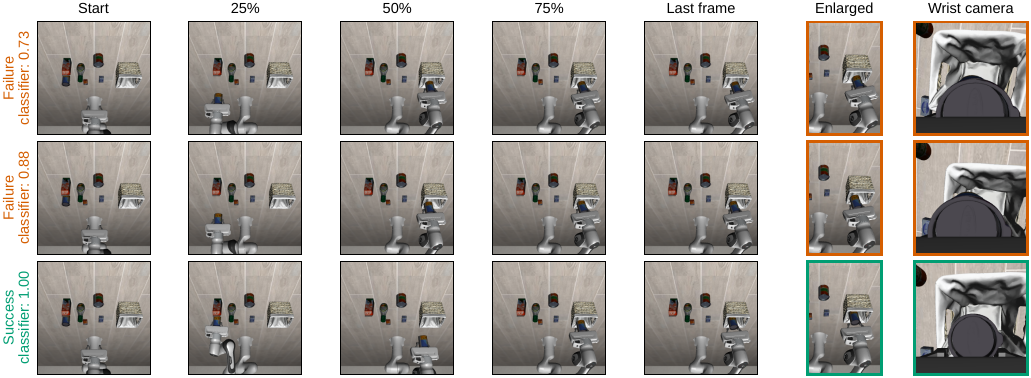}
\caption{\textbf{Failures that look like successes.} Three episodes of one candidate on the same task: two failures that the classifier labels as successes (scores \SLSuppLookScoreA{} and \SLSuppLookScoreB{}) and one success. In the failures the gripper still holds the can at the near edge of the basket when the episode ends; in the success the can is inside it. The final frames differ by a few pixels.}
\label{fig:supp_lookalike}
\end{figure*}

\begin{figure*}[t]
\centering
\includegraphics[width=\textwidth]{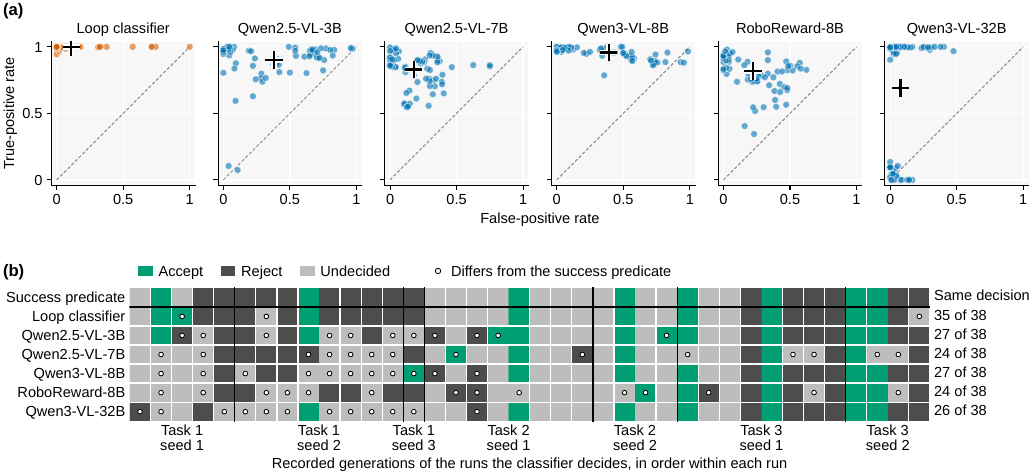}
\caption{\textbf{Six judges on the recorded episodes.} (a) True-positive against false-positive rate, one point per episode set with both outcomes and a cross for all \SLLadderEpisodesPrimary{} episodes pooled. (b) Decision of the test on each judge's verdicts in the \SLLadderPrimaryGens{} recorded generations of the classifier-decided runs, with the decision on the success predicate in the top row. The loop classifier matches it most often.}
\label{fig:supp_ladder}
\end{figure*}

\subsection{Disagreements between the two decisions}
\rhead{Decisions of the two judges.}
\Cref{fig:supp_manip_decisions} shows both decisions for all \SLSuppManipGens{} generations. They differ in \SLSuppManipDiffer{} generations, \SLSuppManipDifferFive{} of them within the first five. In \SLSuppManipUndDec{} of these the classifier's test is undecided at its \SLEvalPerGenInLoopSOne{} pairs while the sealed judge decides, and in \SLSuppManipAccRej{} the classifier accepts a candidate that the sealed judge rejects. The sealed judge stops after a median of \SLSuppManipMedianPairs{} pairs and uses its whole budget in \SLSuppManipAtBudget{} generations. In generations 6 to 8 the classifier kept no candidate (\SLExtensionAcceptedSOne{} of \SLExtensionGenerationsInLoopSOne{}). The two decisions thus differ mostly where the classifier's test cannot decide, and once where it keeps a policy that the sealed judge rejects.

\rhead{Judge, pairs and budget.}
The two tests differ in the judge, in the pairs and in the budget. \Cref{fig:supp_factors} changes one factor at a time. Over the \SLPostBudgetInLoopN{} classifier-decided generations, replacing the classifier by the success predicate on the same pairs changes \SLPostBudgetInLoopJudge{} decisions, replacing the pairs changes \SLPostBudgetInLoopSample{}, and raising the budget changes \SLPostBudgetInLoopBudget{}. The sealed-decided runs give \SLPostBudgetSealedJudge{}, \SLPostBudgetSealedSample{} and \SLPostBudgetSealedBudget{} of \SLPostBudgetSealedN{}. The judge and the budget therefore matter about equally on this testbed. On the same pairs the classifier never accepts a candidate that the predicate rejects. Reading the loop's pairs once with a fixed-sample test on the success predicate (exact McNemar, $\alpha = \SLAlpha$) gives the sealed test's decision in \SLFixedSame{} of \SLFixedGens{} generations, and all \SLFixedAccepted{} candidates it accepts are certified by the sealed test.

\rhead{True success on the classifier's budget.}
A third variant keeps the improver of the classifier-decided runs, with the classifier choosing the training episodes and refreshed in the same way, and changes only the decision: the true success of the same 300 pairs enters one exact McNemar test (two-sided, $\alpha = 0.05$). It runs on the same three tasks with three seeds for five generations, and the sealed test is read in every generation without deciding. It kept \SLPredfixKept{} of \SLPredfixGens{} candidates; the sealed test certified all \SLPredfixKeptCertGain{} as gains, and \SLPredfixGainsLost{} certified gain was rolled back. The decision to keep or roll back equals that of the sealed test in \SLPredfixSameAsSealed{} generations, against \SLInLoopAcceptedWorseSOne{} worse policy kept and \SLFalseRollbacksSOne{} gains lost when the classifier decides. At the classifier's budget of \SLEvalPerGenInLoopSOne{} pairs, true success thus reproduces the sealed decisions, so the wrong decisions of the classifier-decided runs come from the judge and not from the budget.

\subsection{Errors of the classifier}
\rhead{Classifier error over generations.}
\Cref{fig:supp_fpr} follows the classifier's false-positive rate on the failures of each candidate. In the classifier-decided runs its mean is \SLSuppFprFirst{} at the first generation and peaks at \SLSuppFprPeak{} at generation \SLSuppFprPeakGen{}; it exceeds 0.2 in \SLSuppFprAbove{} of all \SLSuppFprGens{} generations. The rate does not rise steadily from one generation to the next. The mean rate of Qwen2.5-VL-3B, which is not trained in the loop, lies between \SLSuppFprVlmLo{} and \SLSuppFprVlmHi{} and is higher than the classifier's in all but one generation of the two panels. The classifier's error thus comes and goes with the failures of each candidate, and a model that is not trained in the loop is not more reliable on them.

\rhead{Accepted failures.}
\Cref{fig:supp_lookalike} shows one generation whose failures look like successes; a video on the project page shows the three episodes in full. Of the candidate's \SLLadderFooledB{} failures, the classifier accepts \SLLadderFooledA{}, and in none of these did the object ever reach the goal region. The classifier frozen before the loop gives the two failures shown scores of \SLSuppLookPreA{} and \SLSuppLookPreB{}, so the new policy found an error that the classifier already had, and retraining did not produce it. A second pattern appears in \SLBlindGenerationsSOne{} generations: after a generation whose collected episodes contain no failure, the retrained classifier accepts failures of the next candidate that the classifier frozen before the loop rejects. The classifier's wrong acceptances thus arise both without retraining and through it.

\begin{figure}[t]
\centering
\includegraphics[width=\linewidth]{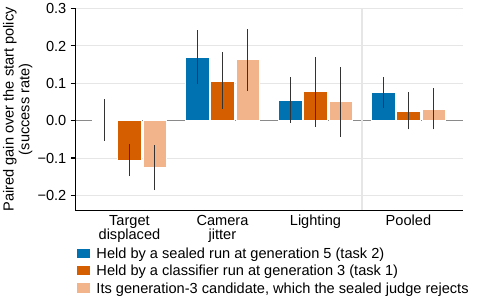}
\caption{\textbf{Three policies on perturbed start conditions.} Paired gain over the start policy by kind of perturbation and pooled (95\% intervals). Two of the three lose when the target is displaced.}
\label{fig:supp_perturbed}
\end{figure}

\begin{figure*}[t]
\centering
\includegraphics[width=\textwidth]{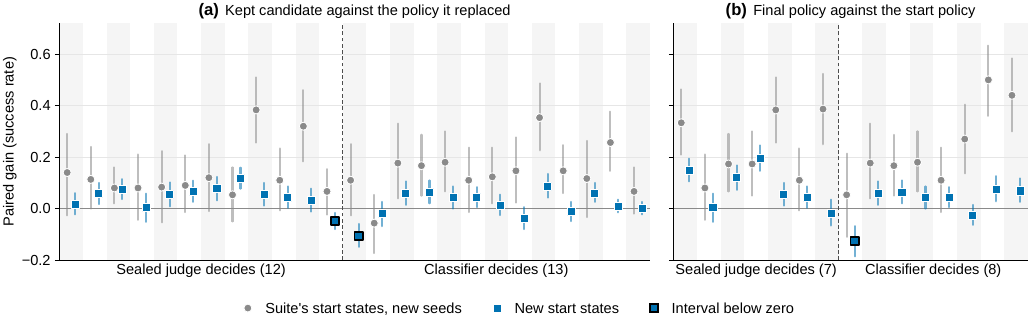}
\caption{\textbf{Kept policies on conditions that no run used.} (a) Each kept candidate against the policy it replaced. (b) The final policy of each run against its start policy; the \SLPostAuditSealedKeptStart{} sealed-decided runs and the \SLPostAuditInLoopKeptStart{} classifier-decided run that kept no candidate end at their start policy and are not drawn. Paired gains with 95\% intervals on the suite's start states with new seeds and on new start states; a black edge marks an interval below zero.}
\label{fig:supp_audit}
\end{figure*}

\subsection{Other judges}
\rhead{Off-the-shelf judges.}
We re-scored the recorded evaluation episodes with four vision-language models and a robot reward model~\cite{ax2601_00675} and ran the test on each judge's verdicts (\cref{fig:supp_ladder}). Each vision-language model sees nine frames of the episode from each of the two cameras and one fixed question, whether the instruction was completed, and its score is the probability of the answer yes; the reward model gives its own progress score. Thresholds are fitted per task on episodes of the start policy. The loop's classifier reaches the success predicate's decision in \SLLadderAgreeJZero{} of \SLLadderPrimaryGens{} generations. The other judges reach it in \SLLadderAgreeJTwo{} (Qwen2.5-VL-3B), \SLLadderAgreeJThree{} (Qwen2.5-VL-7B), \SLLadderAgreeJFour{} (Qwen3-VL-8B), \SLLadderAgreeJFive{} (RoboReward-8B) and \SLLadderAgreeJSix{} (Qwen3-VL-32B). On the \SLJudgesSealedGens{} recorded generations of the sealed-decided runs the classifier again agrees most often, in \SLJudgesSealedAgreeJZero{} against at most \SLJudgesSealedAgreeBestOther{}. Pooled over all episodes the classifier has a true-positive rate of \SLLadderTPRJZero{} at a false-positive rate of \SLLadderFPRJZero{}. Qwen3-VL-32B has the lowest false-positive rate (\SLLadderFPRJSix{}) but a true-positive rate of \SLLadderTPRJSix{}, and it takes \SLLadderSecPerKJSix{}\,s per thousand episodes against \SLLadderSecPerKJFour{}\,s for Qwen3-VL-8B. The off-the-shelf judges remove the two failure patterns only in part. Of the \SLJudgesBlindN{} failures of one generation that follows a refresh without failures, the classifier accepts \SLJudgesBlindJZero{} and every other judge \SLJudgesBlindJSix{}. On the \SLJudgesHardN{} failures of the generation in \cref{fig:supp_lookalike}, the classifier accepts \SLJudgesHardJZero{}; \SLJudgesHardFewer{} of the \SLJudgesOthers{} other judges accept fewer (Qwen2.5-VL-7B \SLJudgesHardJThree{}, Qwen3-VL-32B \SLJudgesHardJSix{}) and \SLJudgesHardMore{} accept more (Qwen2.5-VL-3B \SLJudgesHardJTwo{}, Qwen3-VL-8B \SLJudgesHardJFour{}, RoboReward-8B \SLJudgesHardJFive{}).

\begin{table}[t]
\centering\footnotesize
\setlength{\tabcolsep}{3.5pt}
\begin{tabular}{@{}cc r lr lr@{}}
\toprule
 & & Sealed & \multicolumn{2}{c}{Vision-language judge} & \multicolumn{2}{c}{Success predicate} \\
\cmidrule(lr){4-5}\cmidrule(l){6-7}
Task & Gen. & gain & Decision & Pairs & Decision & Pairs \\
\midrule
1 & 1 & $+0.150$ & Accept & 62 & Accept & 173 \\
 & 2 & $+0.058$$^\dagger$ & Undecided & 300 & Undecided & 300 \\
 & 3 & ${-}0.800$ & Reject & 120 & Reject & 11 \\
 & 4 & ${-}0.438$ & Undecided & 300 & Reject & 29 \\
 & 5 & ${-}0.039$ & Undecided & 300 & Undecided & 300 \\
\midrule
2 & 1 & $+0.029$ & Undecided & 300 & Undecided & 300 \\
 & 2 & $+0.118$$^\dagger$ & Undecided & 300 & Accept & 26 \\
 & 3 & $+0.016$ & Undecided & 300 & Undecided & 300 \\
 & 4 & ${-}0.385$ & Undecided & 300 & Reject & 162 \\
 & 5 & ${-}0.006$ & Undecided & 300 & Undecided & 300 \\
\midrule
3 & 1 & ${-}0.120$ & Reject & 100 & Reject & 210 \\
 & 2 & $+0.168$ & Accept & 111 & Accept & 39 \\
 & 3 & $+0.010$ & Undecided & 300 & Undecided & 300 \\
 & 4 & $+0.095$ & Accept & 253 & Undecided & 300 \\
 & 5 & ${-}0.194$ & Reject & 30 & Reject & 41 \\
\bottomrule
\end{tabular}

\caption{\textbf{A vision-language judge inside the loop.} Decisions of the test on the verdicts of Qwen2.5-VL-3B, which decides, and on the success predicate for the same pairs. $^\dagger$Certified gains that the run rolls back.}
\label{tab:supp_vlm}
\end{table}

\rhead{A vision-language model as the in-loop judge.}
In three further runs, one per task, Qwen2.5-VL-3B replaces the classifier: its verdicts select the training episodes and feed the test, with a fixed threshold (\cref{tab:supp_vlm}). It keeps \SLVLMCertificatesSOne{} of \SLVLMGenerationsSOne{} candidates and no candidate that the sealed judge rejects, and it rolls back \SLVLMFalseRollbacksSOne{} gains that the sealed judge certifies. Its decision equals the predicate's on the same pairs in \SLVLMSameAsPredicateSOne{} generations. A classifier frozen before the loop reaches \SLCZeroSameAsPredicateVLMArmSOne{} on these pairs. The model's test decides within the budget in \SLVLMResolvedSOne{} generations (median stop at \SLVLMMedianStopSOne{} pairs) and the predicate's in \SLPredicateResolvedVLMArmSOne{} (\SLPredicateMedianStopVLMArmSOne{}). Of the \SLVLMPositiveRollbacksSOne{} rolled-back candidates with a positive sealed gain, the predicate accepts \SLVLMJudgeAttributableSOne{} on the same pairs and is undecided on the other \SLVLMBudgetAttributableSOne{}. A vision-language model inside the loop thus keeps no loss, but it gives up certified gains, mostly because its test runs out of budget.

\subsection{Reach of a certified gain}
A certified gain is a statement about the pool the pairs were drawn from. We tested how far it reaches in two ways. First, three policies were compared with their start policy on start conditions with a displaced target, a moved camera or changed lighting (\cref{fig:supp_perturbed}). The policy kept by a sealed-decided run gains \SLPerturbedPoolPiOnePooledSOne{} pooled over the three kinds, and the test accepts it there as well. The policy held by a classifier-decided run at generation 3 and its next candidate, which the sealed judge rejects, gain \SLPerturbedPoolPiThreePooledSOne{} and \SLPerturbedPoolPiFourPooledSOne{}; the test is undecided on both, and both lose when the target is displaced (\SLPerturbedPoolPiThreeTargetShiftSOne{} and \SLPerturbedPoolPiFourTargetShiftSOne{}). Second, all \SLSuppAuditStepSealedN{} candidates kept by the sealed judge and all \SLSuppAuditStepInLoopN{} kept by the classifier were run on \SLPostAuditAPerTask{} conditions per task with the suite's start states and new seeds, and on \SLPostAuditBPerTask{} new start states per task with the target displaced and rotated (\cref{fig:supp_audit}). On the suite's start states the 95\% interval of no kept candidate lies below zero, and the final policies of the nine runs of each variant gain on average \SLPostAuditSealedFinalAMean{} (sealed) and \SLPostAuditInLoopFinalAMean{} (classifier) over their start policies, with a gain of zero for the runs that kept no candidate. On the new start states the gains shrink to \SLPostAuditSealedFinalBMean{} and \SLPostAuditInLoopFinalBMean{}, the interval of one kept candidate of each variant lies below zero, and one classifier-decided run ends \SLPostAuditWorstFinalB{} below its start policy. A certified gain holds for the start states of its pool and carries over to new ones only in part.

\section{Driving}
\label{sec:supp_drive}

\begin{figure*}[t]
\centering
\includegraphics[width=\textwidth]{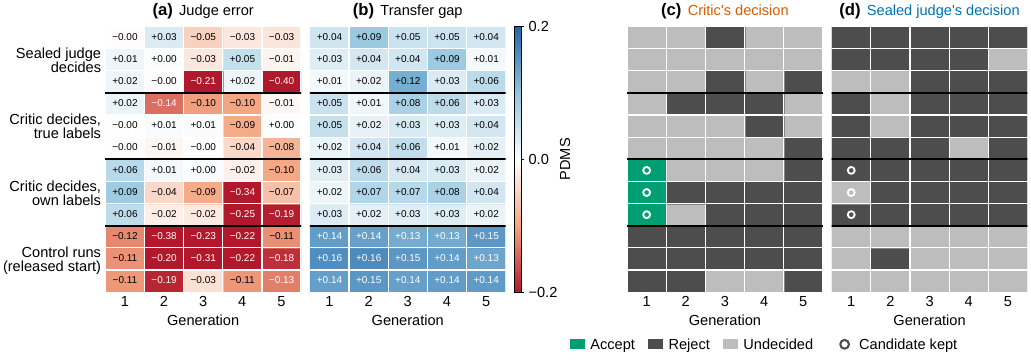}
\caption{\textbf{Every driving run.} Rows are runs, columns are generations. (a) Judge error and (b) transfer gap of each candidate. (c) Decision of the test on the critic's scores and (d) on the sealed judge; a ring marks a candidate that the run kept. The transfer gap is positive in every generation, and the critic keeps three candidates, none of which the sealed judge accepts.}
\label{fig:supp_drive_runs}
\end{figure*}

\subsection{Data, policies and judges}
\rhead{Data and start policy.}
The testbed is NAVSIM~\cite{navsim2024} with DiffusionDrive~\cite{diffusiondrive2025} as the policy. All frames from Boston are excluded from the loop, which sees only Las Vegas, Pittsburgh and Singapore. The main start policy is trained from scratch on \SLImproverFramesSThree{} training-city frames for 20 epochs, so that it can still improve. It relies on its camera: with the image replaced by a black or a gray frame, its PDMS on \SLExtraCamFrames{} Boston frames changes by \SLExtraCamBlack{} and \SLExtraCamGrey{}. We chose this policy among \SLSubsetConfigs{} training configurations as the one with the highest PDMS on \SLSubsetSelFrames{} Boston frames (\SLSubsetStartPDMS{}). These frames are \SLSubsetSelShare{} of the sealed pool. As a check, we ran each generation's sealed test again on the other \SLSubsetFrames{} frames, from \SLSubsetLogs{} logs that played no part in the choice. The decision is the same for \SLSubsetSame{} of the \SLSubsetCandidates{} candidates of the nine main runs; \SLSubsetChanged{} rejections become undecided and no candidate is accepted. The gain is negative for \SLSubsetNegative{} candidates, and its 95\% interval over logs lies below zero for \SLSubsetBelowZero{} and above zero for none. The three candidates that the critic kept change the score on these frames by \SLSubsetKeptA{}, \SLSubsetKeptB{} and \SLSubsetKeptC{}. Choosing the start policy on part of the sealed pool therefore does not change the outcome of the runs.

\rhead{Improver and critic.}
In every generation the policy draws \SLProposalsSThree{} trajectory proposals on \SLImproverFramesSThree{} new unlabeled frames, the critic picks one proposal per frame, and the policy is trained for five epochs on these picks, mixed one to one with expert frames. The fifth of the frames on which the critic's pick scores furthest above the policy's own first choice is duplicated. The in-loop judge is this critic, a network that predicts PDMS from frozen encoder features and a trajectory, fitted before the loop on \SLImproverFramesSThree{} frames with the true PDMS of their proposals. It is refreshed every generation in one of two ways. With \emph{true labels}, 2{,}000 of the generation's frames receive the true PDMS of their proposals and the critic is fitted again on all labels so far. With \emph{own labels}, the critic is trained further to prefer its own pick over the other proposals of each frame.

\rhead{Judges and runs.}
The critic's test reads its scores for both policies on \SLInLoopFramesSThree{} training-city frames disjoint from all training data. Their logs are disjoint from the logs of the improver's frames and of the critic's labels; the start policy was trained on other frames of the same logs. The sealed judge is the true PDMS on \SLPoolFramesSThree{} Boston frames from 280 logs. Gains are means per log; scores of single policies are means over frames. Three seeds of three variants (sealed judge decides, critic decides with either refresh) give nine main runs of five generations. Three control runs, one per variant, start from the released DiffusionDrive checkpoint and are judged on 3{,}731 Boston frames from 51 logs.

\rhead{Learning rate.}
To set the improver's learning rate, we trained the start policy on its own first choice in place of the critic's pick. At learning rates of $3{\times}10^{-5}$, $1{\times}10^{-4}$ and $6{\times}10^{-4}$ its PDMS on \SLExtraNullFrames{} frames changes by \SLExtraNullLow{}, \SLExtraNullMid{} and \SLExtraNullHigh{}, and the test accepts none of the three. All runs use the largest rate.

\begin{table}[t]
\centering\footnotesize
\setlength{\tabcolsep}{2.6pt}
\begin{tabular}{@{}>{\columncolor{white}[0pt][\tabcolsep]}l r ccc >{\columncolor{white}[\tabcolsep][0pt]}r@{}}
\toprule
Frames & $n$ & Rank corr. & Top-1 & FPR & Pick $-$ own \\
\midrule
\rowcolor{black!8}\multicolumn{6}{@{}>{\columncolor{white}[0pt][0pt]}l@{}}{\emph{Main runs (start policy trained from scratch)}} \\
Loop's own frames & 5{,}000 & 0.74 & 0.67 & 0.16 & $+0.010 \pm 0.003$ \\
Unseen logs & 2{,}000 & 0.73 & 0.68 & 0.18 & $+0.006 \pm 0.005$ \\
Boston (sealed pool) & 16{,}685 & 0.40 & 0.25 & 0.38 & ${-}0.042 \pm 0.004$ \\
\midrule
\rowcolor{black!8}\multicolumn{6}{@{}>{\columncolor{white}[0pt][0pt]}l@{}}{\emph{Control runs (released checkpoint)}} \\
Loop's own frames & 5{,}000 & 0.75 & 0.73 & 0.18 & $+0.001 \pm 0.002$ \\
Unseen logs & 2{,}000 & 0.75 & 0.71 & 0.17 & $+0.002 \pm 0.003$ \\
Boston (sealed pool) & 3{,}731 & 0.48 & 0.39 & 0.47 & ${-}0.073 \pm 0.008$ \\
\bottomrule
\end{tabular}

\caption{\textbf{The critic before the first generation.} Rank correlation of its scores with the true PDMS, share of frames where its pick has the highest true PDMS (top-1), share of zero-PDMS proposals it scores above its threshold (FPR), and true PDMS of its pick minus that of the policy's first choice ($\pm$ standard error over logs). Unseen logs are training-city logs outside the loop.}
\label{tab:supp_critic}
\end{table}

\begin{figure*}[t]
\centering
\includegraphics[width=\textwidth]{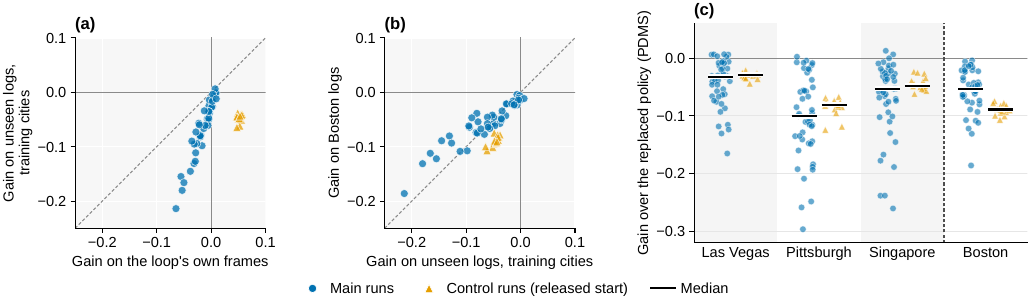}
\caption{\textbf{Where the gain is lost.} True gain of every candidate over the policy it would replace. (a) On unseen logs of the training cities against the loop's own evaluation frames. (b) On Boston logs against unseen logs of the training cities. (c) By city, with medians. Most of the loss appears on unseen logs of the training cities.}
\label{fig:supp_transfer}
\end{figure*}

\begin{figure*}[t]
\centering
\includegraphics[width=\textwidth]{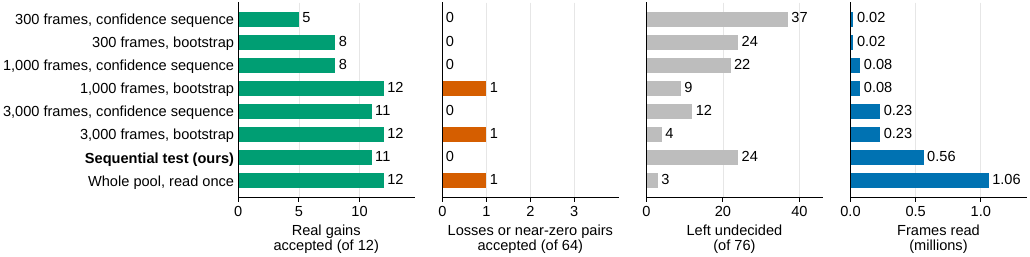}
\caption{\textbf{Acceptance rules on \SLSuppRulesPairs{} pairs with known gains.} For each rule: real gains accepted, losses or pairs near zero accepted, pairs left undecided and frames read. Fixed-size rules read a sample once, with the confidence sequence or a 95\% bootstrap over logs. The sequential test accepts no loss and no pair near zero and reads about half of what the whole pool costs.}
\label{fig:supp_rules}
\end{figure*}

\subsection{The critic and the judge gap}
\rhead{The critic before the first generation.}
\Cref{tab:supp_critic} compares the critic with the true PDMS before any policy has been trained on its picks. On the loop's own frames and on unseen logs of the training cities it ranks proposals well, and its pick is slightly better than the policy's own choice (\SLCriticSelTrainSThree{} PDMS). In Boston the rank correlation falls by almost half, the false-positive rate rises from \SLCriticFPRTrainSThree{} to \SLCriticFPRHeldSThree{}, and its pick is worse than the policy's own choice (\SLCriticSelHeldSThree{}). The control runs show the same pattern (\SLCriticSelTrainCtlSThree{} against \SLCriticSelHeldCtlSThree{}). The critic is therefore already unreliable outside its training cities before the loop starts.

\rhead{Results by run.}
\Cref{fig:supp_drive_runs} gives the two terms of the judge gap and both decisions for all \SLSuppDriveMainGens{} main and \SLSuppDriveCtlGens{} control generations. The transfer gap is positive in all \SLSuppDriveGapPos{} of them, \SLSuppDriveGapMain{} PDMS on average in the main runs and \SLSuppDriveGapCtl{} in the control runs. The judge error averages \SLSuppDriveErrMain{} and \SLSuppDriveErrCtl{}: after the first generation the critic mostly reports less gain than the candidate has on the loop's own frames. In the control runs the judge error is \SLControlJudgeErrorRangeSThree{} and the transfer gap \SLControlTransferRangeSThree{} across generations: the critic under-reports in every one, and the whole positive gap is transfer. It keeps \SLSuppDriveCriticAcc{} candidates, all at the first generation of the runs that refresh it on its own labels. The sealed judge accepts \SLSuppDriveSealedAcc{} of the \SLSuppDriveMainGens{} main candidates, rejects \SLSuppDriveSealedRej{} and is undecided on \SLSuppDriveSealedUnd{}. In the control runs the critic rejects \SLSuppDriveCtlCriticRej{} of \SLSuppDriveCtlGens{} candidates, and the sealed judge rejects \SLSuppDriveCtlSealedRej{} and is undecided on \SLSuppDriveCtlSealedUnd{}, since its pool holds only 51 logs. Refreshing does not repair the critic outside its cities: in the control runs the rank correlation of the refreshed critics with the true score in Boston is \SLControlRefreshedSpearmanRangeSThree{}, below the value before the loop (\cref{tab:supp_critic}).

\begin{table*}[t]
\centering\small
\setlength{\tabcolsep}{6pt}
\setbox0\hbox{\begin{tabular}{@{}>{\columncolor{white}[0pt][\tabcolsep]}lccccc>{\columncolor{white}[\tabcolsep][0pt]}c@{}}
\toprule
 & Gains & \multicolumn{2}{c}{Other pairs accepted} & Frames & \multicolumn{2}{c}{Accepted, no-gain pools} \\
\cmidrule(lr){3-4}\cmidrule(lr){6-7}
Rule & accepted & Losses & Near zero & read & Mean & Worst pair \\
\midrule
\ours Sequential test (ours) & 11.2 & \good 0 & \good 0 & 53\% & \good 0.0\% & \good 0.0\% \\
50 logs once, bootstrap & 10.8 & \good 0 & \bad 0.17 & 22\% & \bad 0.8\% & \bad 2.5\% \\
50 logs once, confidence sequence & 1.1 & \good 0 & \good 0 & 22\% & \good 0.0\% & \good 0.0\% \\
17 logs once, bootstrap & 8.4 & \bad 0.01 & \bad 0.22 & 8\% & \bad 2.2\% & \bad 4.2\% \\
5 logs once, bootstrap & 6.1 & \bad 0.25 & \bad 0.45 & 2\% & \bad 6.3\% & \bad 10.7\% \\
Bootstrap after every 10 logs & 12.0 & \bad 0.06 & \bad 1.16 & 13\% & \bad 5.9\% & \bad 11.4\% \\
\bottomrule
\end{tabular}
}\addtolength{\tabcolsep}{\dimexpr(\textwidth-\wd0)/12\relax}
\begin{tabular}{@{}>{\columncolor{white}[0pt][\tabcolsep]}lccccc>{\columncolor{white}[\tabcolsep][0pt]}c@{}}
\toprule
 & Gains & \multicolumn{2}{c}{Other pairs accepted} & Frames & \multicolumn{2}{c}{Accepted, no-gain pools} \\
\cmidrule(lr){3-4}\cmidrule(lr){6-7}
Rule & accepted & Losses & Near zero & read & Mean & Worst pair \\
\midrule
\ours Sequential test (ours) & 11.2 & \good 0 & \good 0 & 53\% & \good 0.0\% & \good 0.0\% \\
50 logs once, bootstrap & 10.8 & \good 0 & \bad 0.17 & 22\% & \bad 0.8\% & \bad 2.5\% \\
50 logs once, confidence sequence & 1.1 & \good 0 & \good 0 & 22\% & \good 0.0\% & \good 0.0\% \\
17 logs once, bootstrap & 8.4 & \bad 0.01 & \bad 0.22 & 8\% & \bad 2.2\% & \bad 4.2\% \\
5 logs once, bootstrap & 6.1 & \bad 0.25 & \bad 0.45 & 2\% & \bad 6.3\% & \bad 10.7\% \\
Bootstrap after every 10 logs & 12.0 & \bad 0.06 & \bad 1.16 & 13\% & \bad 5.9\% & \bad 11.4\% \\
\bottomrule
\end{tabular}

\caption{\textbf{Acceptance rules on one unit.} Every rule reads whole logs of each of the \SLFairPairs{} pairs in \SLFairOrders{} random orders; entries are means per pass over the pairs. Frames read are a share of the whole pools. Last two columns: share accepted of the \SLFairNullPools{} pools without a gain, built by swapping the two scores in a random half of the logs and keeping the pools whose mean gain is not positive.}
\label{tab:supp_rules_fair}
\end{table*}

\begin{table}[t]
\centering\small
\setlength{\tabcolsep}{2pt}
\begin{tabular}{@{}ccc c cc>{\columncolor{white}[\tabcolsep][0pt]}c@{}}
\toprule
 & & Kept & Sealed & \multicolumn{3}{c}{Decision} \\
\cmidrule(l){5-7}
Seed & Gen. & before & gain & Sealed & Own frames & Critic \\
\midrule
0 & 1 & 0 & $+0.031$ & \good Accept & \good Accept & Undecided \\
0 & 2 & 1 & $+0.013$ & Undecided & \good Accept & Undecided \\
0 & 2 & 0 & $+0.043$ & \good Accept & \good Accept & Undecided \\
0 & 3 & 2 & $+0.009$ & Undecided & \good Accept & Undecided \\
0 & 3 & 1 & $+0.017$ & Undecided & \good Accept & Undecided \\
0 & 3 & 0 & $+0.041$ & \good Accept & \good Accept & Undecided \\
0 & 4 & 3 & $-0.004$ & Undecided & Undecided & Undecided \\
0 & 4 & 1 & $+0.010$ & Undecided & Undecided & Undecided \\
0 & 4 & 0 & $+0.029$ & \good Accept & \good Accept & Undecided \\
0 & 5 & 3 & $-0.004$ & Undecided & Undecided & Undecided \\
0 & 5 & 1 & $+0.011$ & Undecided & Undecided & Undecided \\
0 & 5 & 0 & $+0.036$ & \good Accept & \good Accept & Undecided \\
\midrule
1 & 1 & 0 & $+0.042$ & \good Accept & \good Accept & Undecided \\
1 & 2 & 1 & $+0.007$ & Undecided & \good Accept & Undecided \\
1 & 2 & 0 & $+0.039$ & \good Accept & \good Accept & Undecided \\
1 & 3 & 2 & $-0.001$ & Undecided & Undecided & Undecided \\
1 & 3 & 1 & $+0.001$ & Undecided & \good Accept & Undecided \\
1 & 3 & 0 & $+0.034$ & \good Accept & \good Accept & Undecided \\
1 & 4 & 2 & $+0.005$ & Undecided & Undecided & Undecided \\
1 & 4 & 1 & $+0.008$ & Undecided & \good Accept & Undecided \\
1 & 4 & 0 & $+0.039$ & \good Accept & \good Accept & Undecided \\
1 & 5 & 2 & $+0.005$ & Undecided & Undecided & Undecided \\
1 & 5 & 1 & $+0.004$ & Undecided & \good Accept & Undecided \\
1 & 5 & 0 & $+0.039$ & \good Accept & \good Accept & Undecided \\
\midrule
2 & 1 & 0 & $+0.034$ & \good Accept & \good Accept & \good Accept \\
2 & 2 & 1 & $+0.008$ & Undecided & \good Accept & Undecided \\
2 & 3 & 2 & $+0.007$ & Undecided & Undecided & Undecided \\
2 & 3 & 1 & $+0.010$ & Undecided & \good Accept & Undecided \\
2 & 4 & 2 & $+0.009$ & Undecided & Undecided & Undecided \\
2 & 4 & 1 & $+0.017$ & Undecided & \good Accept & Undecided \\
2 & 5 & 2 & $+0.007$ & Undecided & \good Accept & Undecided \\
2 & 5 & 1 & $+0.016$ & Undecided & \good Accept & Undecided \\
\bottomrule
\end{tabular}
\caption{\textbf{Every candidate of the loop that improves.} Number of earlier candidates kept on its history, sealed gain over the policy it would replace (PDMS) and the decision under each judge.}
\label{tab:supp_work}
\end{table}

\subsection{Origin of the transfer gap}
\rhead{Training on the best proposals.}
We repeated the first generation with targets chosen by the true score in place of the critic. On each of \SLAddOracleFrames{} training frames the target is the proposal with the highest PDMS, \SLAddOracleTargetPDMS{} on average against \SLAddOraclePolicyPDMS{} for the policy's own first choice. At learning rates of $2{\times}10^{-5}$, $1{\times}10^{-4}$ and $6{\times}10^{-4}$ the candidate changes PDMS on the \SLAddOracleBostonFrames{} Boston frames by \SLAddOracleGainLow{} \SLAddOracleCILow{}, \SLAddOracleGainMid{} \SLAddOracleCIMid{} and \SLAddOracleGainHigh{} \SLAddOracleCIHigh{} (95\% bootstrap intervals over logs). Better targets alone thus do not give the candidate a gain in Boston.

\rhead{Loss on unseen logs of the training cities.}
The sealed pool differs from the loop's frames in two ways: its logs are new and its city is new. To separate the two, we scored every candidate on \SLSuppTransferFrames{} frames from logs of the training cities that no run used (\cref{fig:supp_transfer}). \SLPostTransferMainNeg{} of \SLPostTransferMainN{} main candidates lose on these frames, with a median of \SLPostTransferMainPThreeMed{} PDMS against \SLPostTransferMainBostonMed{} in Boston. Reweighting these frames to the city mix of the loop's own evaluation frames leaves the loss in place: the median is \SLPostTransferMainMixMed{} for the main candidates (\SLPostTransferMainMixRange{}) and \SLPostTransferCtlMixMed{} for the control candidates (\SLPostTransferCtlMixRange{}), so the loss on unseen logs does not come from a different share of the three cities. The control candidates gain \SLPostTransferCtlVinMed{} on the loop's own evaluation frames and lose \SLPostTransferCtlPThreeMed{} on unseen logs of the same cities and \SLPostTransferCtlBostonMed{} in Boston, so \SLPostTransferCtlShare{} of the drop is present before the city changes. The median gain of the main candidates is \SLSuppCityVegas{} in Las Vegas, \SLSuppCityPitt{} in Pittsburgh, \SLSuppCitySing{} in Singapore and \SLSuppCityBoston{} in Boston. A critic without error would not have helped the control runs: the test on the true PDMS of the loop's own evaluation frames accepts \SLVinReplayFrameAcceptCtlSThree{} control generations per frame and \SLVinReplayBlockAcceptCtlSThree{} per log, and all of them lose in Boston. On the main runs the same test accepts \SLVinReplayMainSThree{}. Applied to the unseen logs of the training cities, the same test keeps none of the \SLPThreeReplayN{} candidates and rejects \SLPThreeReplayRej{}. Most of the transfer gap thus opens between logs of the same cities, where no judge that reads the loop's own frames can see it.

\begin{table*}[t]
\centering\small
\setlength{\tabcolsep}{6pt}
\setbox0\hbox{\begin{tabular}{@{}lccccc@{}}
\toprule
 & & & & Scenes penalized only & PDMS change, \\
Term & IL & RL & RL $-$ IL & for IL / for RL & this term alone \\
\midrule
No at-fault collision (NC) & 0.980 & 0.981 & \ensuremath{+0.001} & 101 / 88 & \ensuremath{-0.004} \\
Drivable-area compliance (DAC) & 0.946 & 0.976 & \ensuremath{+0.030} & 440 / 72 & \ensuremath{+0.015} \\
Time to collision within bound (TTC) & 0.941 & 0.951 & \ensuremath{+0.010} & 337 / 216 & \ensuremath{+0.002} \\
Ego progress (EP) & 0.804 & 0.858 & \ensuremath{+0.054} &  & \ensuremath{+0.006} \\
Comfort (C) & 1.000 & 1.000 & \ensuremath{+0.000} & 5 / 0 & \ensuremath{+0.000} \\
\midrule
PDMS & 0.861 & 0.904 & \ensuremath{+0.043} & & \\
\midrule
Driving-direction compliance (DDC)$^\ast$ & 0.981 & 0.967 & \ensuremath{-0.014} & 31 / 221 &  \\
\bottomrule
\end{tabular}
}\addtolength{\tabcolsep}{\dimexpr(\textwidth-\wd0)/10\relax}
\begin{tabular}{@{}lccccc@{}}
\toprule
 & & & & Scenes penalized only & PDMS change, \\
Term & IL & RL & RL $-$ IL & for IL / for RL & this term alone \\
\midrule
No at-fault collision (NC) & 0.980 & 0.981 & \ensuremath{+0.001} & 101 / 88 & \ensuremath{-0.004} \\
Drivable-area compliance (DAC) & 0.946 & 0.976 & \ensuremath{+0.030} & 440 / 72 & \ensuremath{+0.015} \\
Time to collision within bound (TTC) & 0.941 & 0.951 & \ensuremath{+0.010} & 337 / 216 & \ensuremath{+0.002} \\
Ego progress (EP) & 0.804 & 0.858 & \ensuremath{+0.054} &  & \ensuremath{+0.006} \\
Comfort (C) & 1.000 & 1.000 & \ensuremath{+0.000} & 5 / 0 & \ensuremath{+0.000} \\
\midrule
PDMS & 0.861 & 0.904 & \ensuremath{+0.043} & & \\
\midrule
Driving-direction compliance (DDC)$^\ast$ & 0.981 & 0.967 & \ensuremath{-0.014} & 31 / 221 &  \\
\bottomrule
\end{tabular}

\caption{\textbf{PDMS terms of the two versions.} Means over the \SLVehicleSubScenes{} navtest scenes scored for both. Penalized: the term is below one. Last column: change in mean PDMS when only this term takes the RL values. $^\ast$Not part of PDMS.}
\label{tab:supp_vehicle_subscores}
\end{table*}

\begin{figure*}[t]
\centering
\includegraphics[width=\textwidth]{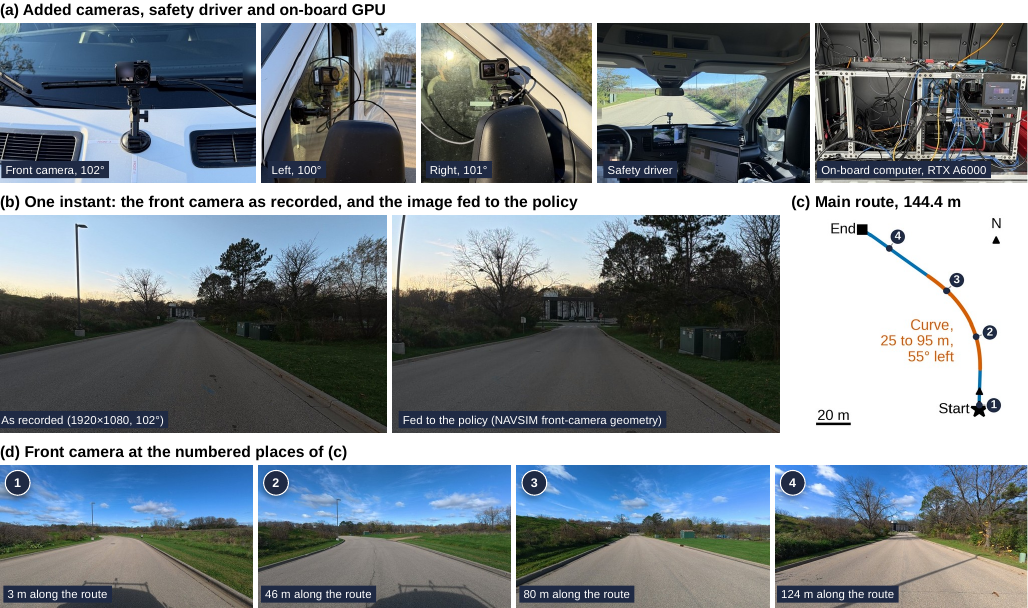}
\caption{\textbf{Vehicle setup and route.} (a) The three added cameras, the safety driver and the on-board GPU; both versions plan from the front camera only. (b) One instant as recorded by the front camera and as fed to the policy, reprojected to the geometry of the NAVSIM front camera. (c) Main route with its left curve. (d) Front camera at the numbered places of (c).}
\label{fig:supp_vehicle_cameras}
\end{figure*}

\subsection{The sequential test on known gains}
\rhead{Pairs with known gains.}
To test the sequential test on real gains, we ran it on \SLPostCtlPairs{} pairs of policies whose gain is known from the whole pool: the \SLPostCtlLoopPairs{} candidates, \SLPostCtlWeakPairs{} pairs among \SLPostCtlWeakStarts{} start policies trained from scratch with different amounts of data, and the main start policy against the released checkpoint. A pair counts as a real gain or loss when the 95\% bootstrap interval over the logs of the whole pool lies above or below zero. This gives \SLPostCtlGains{} gains, \SLPostCtlLosses{} losses and \SLPostCtlNearZero{} pairs near zero (\cref{fig:supp_rules}). The sequential test accepts \SLPostCtlSeqGainsAcc{} of the \SLPostCtlGains{} gains and no other pair. The one gain it misses (\SLPostCtlSeqMissed{} PDMS) stays undecided with the whole pool read, and the accepted ones need \SLPostCtlSeqAccFramesRange{} frames each. Over the \SLPostCtlPairs{} pairs it reads \SLPostCtlSeqFrames{} frames, counting every frame of the logs it opened. In the nine main runs the sequential test reads \SLSeqFramesMainSThree{} of the \SLSeqPoolFramesMainSThree{} frames that scoring the whole pool in every generation would need (\SLSeqShareMainSThree{}). It decides \SLSeqDecidedMainSThree{} generations after a median of \SLSeqStopMedianMainSThree{} frames and leaves \SLSeqUndecidedMainSThree{} undecided. The test thus accepts nearly every real gain and nothing else, at about half the cost of scoring the whole pool.

\rhead{Sequential test and fixed sample.}
The sequential test adds an error rate that holds wherever the reading stops. \Cref{tab:supp_rules_fair} puts the rules on one unit. Every rule reads whole logs in a random order, the gain of a pair is the mean difference per log, and each pair is read in \SLFairOrders{} orders. By the per-log mean, \SLFairGains{} of the \SLFairPairs{} pairs are gains, \SLFairLosses{} are losses and \SLFairNearZero{} lie near zero. A bootstrap over 50 logs read once accepts \SLFairBootFiftyGains{} of the \SLFairGains{} gains on average with \SLFairBootFiftyShare{} of the frames. The sequential test accepts \SLFairSeqGains{} with \SLFairSeqShare{}, \SLFairSeqOverBootFifty{} times as many frames, and neither rule accepts a loss. The sequential test differs in that it may read on. At 50 logs the confidence sequence has accepted \SLFairCSFiftyGains{} gains and continues. The bootstrap read after every \SLFairPeekEvery{} logs until its interval leaves zero accepts \SLFairPeekNearAcc{} pairs near zero per pass, one of them in \SLFairPeekNearMaxPair{} of the orders. We also built pools without a gain by swapping the two scores in a random half of the logs and keeping the \SLFairNullPools{} pools whose mean gain is not positive. The bootstrap read after every \SLFairPeekEvery{} logs accepts \SLFairPeekNullFA{} of them and up to \SLFairPeekNullFAMax{} for one pair, against a nominal 2.5\%. Read once over 50 logs it accepts \SLFairBootFiftyNullFA{}, and the sequential test accepts none. With a budget fixed in advance, a bootstrap over logs is the cheaper rule. The sequential test is the rule whose error rate holds when the amount of evaluation is decided while reading.

\subsection{A loop built to improve}
Here the start policy is DiffusionDrive trained on 5{,}000 frames, and each generation imitates the logged expert on 20{,}000 new frames, so real gains are expected. Each of the \SLWorkSeeds{} seeds runs \SLWorkGens{} generations under three deciding judges: the learned critic with the sequential test, the true score on the loop's own frames with one paired $t$ test, and the sealed judge with the sequential test (\cref{tab:supp_work}). Of the \SLWorkCands{} candidates, \SLWorkFirst{} are a first step from the start policy, reached over the three seeds and different roll-back histories. The sealed test accepted all of them; the learned critic accepted \SLWorkFirstCriticAcc{} and left the start policy in place in \SLWorkCriticStuckSeeds{} of three seeds. The other \SLWorkLater{} build on an accepted policy; the sealed test left all of them undecided, and the paired test on the loop's own frames accepted \SLWorkLaterVinAcc{} of them (\SLWorkVinAcc{} of all \SLWorkCands{}). The sealed test read \SLWorkSeqShare{} of the pool on average. The gains transfer in this loop, so the loss under the critic comes from its error, and the sealed test recovers \SLWorkSealedRise{} of the \SLWorkVinRise{} PDMS that the paired test on true scores reaches.

\section{Real-world field experiment}
\label{sec:supp_vehicle}

\subsection{The two versions before the drives}
\rhead{The two versions in simulation.}
We scored both versions of ReCogDrive~\cite{ax2506_08052} on the NAVSIM navtest split with one pipeline. On the \SLVehicleNavsimTokensSFour{} scenes scored for both, the imitation version (IL) reaches \SLVehicleNavsimILSFour{} PDMS and the reinforced version (RL) \SLVehicleNavsimRLSFour{}, a paired difference of \SLVehicleNavsimPairedSFour{}. \Cref{tab:supp_vehicle_subscores} splits the difference by the terms of PDMS. RL gains by staying in the drivable area (\SLVehicleSubDACIL{} to \SLVehicleSubDACRL{}; \SLVehicleSubDACOnlyIL{} scenes penalized only for IL against \SLVehicleSubDACOnlyRL{} only for RL) and by progress (\SLVehicleSubEPIL{} to \SLVehicleSubEPRL{}). At-fault collisions change little (\SLVehicleSubNCOnlyIL{} against \SLVehicleSubNCOnlyRL{} scenes), and driving-direction compliance, which PDMS leaves out, falls from \SLVehicleSubDDCIL{} to \SLVehicleSubDDCRL{}. The routes with a curve therefore test drivable-area compliance, the term that contributes most to the simulator's gain.

\rhead{Plans on the vehicle's camera.}
We ran both versions on \SLExtraReplayWindows{} camera windows recorded on the vehicle under a human driver. Their planned trajectories differ by \SLExtraReplayDiff{}\,m on average over the 4\,s horizon, and by more than 0.5\,m in \SLExtraReplayShare{} of the windows. Repeated sampling of one version moves its plan by \SLExtraReplaySpreadIL{}\,m (IL) and \SLExtraReplaySpreadRL{}\,m (RL). The two versions thus differ on real camera input by about six times their sampling spread.

\subsection{Vehicle and procedure}
\rhead{Full-scale vehicle and software.}
The vehicle is a full-scale electric van with a drive-by-wire kit, an RTK-corrected inertial navigation system, three wide-angle cameras and an on-board NVIDIA RTX A6000 (\cref{fig:supp_vehicle_cameras}). The two versions share the vision-language backbone and differ only in the weights of the diffusion planner. Each plans an eight-point trajectory over 4\,s from the front camera with five diffusion steps, once per \SLVehPlanPeriod{}\,s. The camera calibration, the trajectory tracker and the speed limit of 4\,m/s are the same for both.

\rhead{Routes and scoring.}
The main route (\cref{fig:supp_vehicle_cameras}c) is 144.4\,m long: about 25\,m straight, a left curve of about 55$^\circ$ between about 25 and 95\,m, and about 50\,m straight. A straight route (60.0\,m) and a route with a right curve (143.1\,m) lie on the same road. Every drive starts with the rear axle within 3\,m of the route start, in daylight on a dry road, with a safety driver who keeps a foot over the brake. A drive ends at the first intervention (a brake, steering or throttle override, a timeout or a stall) or at the end of the route. Its score is $\min(D, L)/L$, where $D$ is the distance along the route at the first intervention and $L$ the route length.

\begin{table}[t]
\centering\footnotesize
\setlength{\tabcolsep}{2pt}
\setbox0\hbox{\begin{tabular}{@{}llllll@{}}
\toprule
Route & Pair & Order & IL & RL & RL $-$ IL (route fraction) \\
\midrule
Left & 1 & IL first & 43\,m, brake & 55\,m, brake & \ensuremath{+0.082} \\
 & 2 & RL first & 30\,m, brake & 57\,m, brake & \ensuremath{+0.186} \\
 & 3 & RL first & 39\,m, brake & 49\,m, brake & \ensuremath{+0.069} \\
 & 4 & IL first & 45\,m, brake & 47\,m, brake & \ensuremath{+0.008} \\
 & 5 & RL first & 45\,m, brake & 48\,m, brake & \ensuremath{+0.018} \\
 & 6 & IL first & 44\,m, brake & 54\,m, steer & \ensuremath{+0.069} \\
\midrule
Straight & 1 & IL first & completed & completed & \ensuremath{+0.000} \\
 & 2 & RL first & completed & completed & \ensuremath{+0.000} \\
 & 3 & IL first & completed & completed & \ensuremath{+0.000} \\
\midrule
Right & 1 & RL first & 83\,m, brake & 98\,m, steer & \ensuremath{+0.106} \\
\bottomrule
\end{tabular}
}\addtolength{\tabcolsep}{\dimexpr(\linewidth-\wd0)/10\relax}
\begin{tabular}{@{}llllll@{}}
\toprule
Route & Pair & Order & IL & RL & RL $-$ IL (route fraction) \\
\midrule
Left & 1 & IL first & 43\,m, brake & 55\,m, brake & \ensuremath{+0.082} \\
 & 2 & RL first & 30\,m, brake & 57\,m, brake & \ensuremath{+0.186} \\
 & 3 & RL first & 39\,m, brake & 49\,m, brake & \ensuremath{+0.069} \\
 & 4 & IL first & 45\,m, brake & 47\,m, brake & \ensuremath{+0.008} \\
 & 5 & RL first & 45\,m, brake & 48\,m, brake & \ensuremath{+0.018} \\
 & 6 & IL first & 44\,m, brake & 54\,m, steer & \ensuremath{+0.069} \\
\midrule
Straight & 1 & IL first & completed & completed & \ensuremath{+0.000} \\
 & 2 & RL first & completed & completed & \ensuremath{+0.000} \\
 & 3 & IL first & completed & completed & \ensuremath{+0.000} \\
\midrule
Right & 1 & RL first & 83\,m, brake & 98\,m, steer & \ensuremath{+0.106} \\
\bottomrule
\end{tabular}

\caption{\textbf{Every scored pair of the field experiment.} IL and RL: distance along the route at the first intervention and its kind (brake or steering override), or completed.}
\label{tab:supp_vehicle_records}
\end{table}

\begin{figure}[t]
\centering
\includegraphics[width=\linewidth]{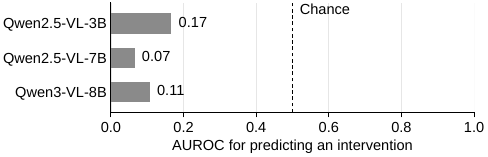}
\caption{\textbf{Vision-language judges on drive videos.} AUROC for predicting which of the \SLVehScored{} scored drives needed an intervention.}
\label{fig:supp_vehicle_judges}
\end{figure}

\rhead{Order and blinding.}
The two versions are driven in pairs from the same start. Pairs come in blocks of two, and the order within each block is random. The safety driver and the operator saw the versions only under two blind labels. After every drive the safety driver guessed the version. The guesses were right in \SLVehGuessLeft{} drives on the main route, \SLVehGuessStraight{} on the straight route (\SLVehGuessStraightDK{} answers of ``don't know'') and \SLVehGuessRight{} on the right curve.

\begin{figure*}[t]
\centering
\includegraphics[width=\textwidth]{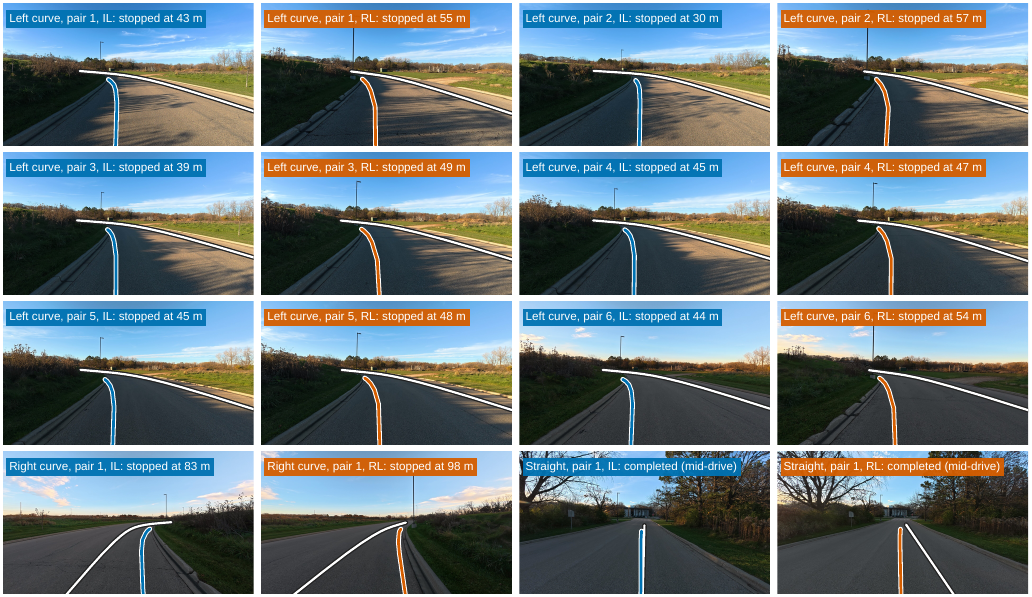}
\caption{\textbf{Front camera at the intervention.} Last planning frame before the safety driver took over, for every pair on the two curves, with the planned trajectory over 4\,s (blue: IL, orange: RL) and the route (white) drawn on the road. The last two frames show the first pair on the straight route in mid-drive.}
\label{fig:supp_vehicle_frames}
\end{figure*}

\subsection{Results}
\rhead{Results by pair.}
\Cref{tab:supp_vehicle_records} lists every scored pair: \SLVehLeftPairs{} on the main route, \SLVehStraightPairs{} on the straight route and \SLVehRightPairs{} on the right curve. No drive completed the main route (\SLVehLeftCompleted{}). The safety driver stopped every drive between \SLVehLeftFailMin{} and \SLVehLeftFailMax{}\,m, in the first half of the curve, after \SLVehLeftMeanIL{}\,m on average for IL and \SLVehLeftMeanRL{}\,m for RL. RL went further in \SLVehLeftRLFurther{} pairs (exact sign test, one-sided $p = \SLVehLeftSignP{}$), by \SLVehLeftMeanDiff{} of the route on average, and the confidence sequence after \SLVehLeftPairs{} pairs is \SLVehLeftCS{}, so the sequential test is undecided. Every drive on the straight route completed it (\SLVehStraightCompleted{}). On the right curve IL was stopped at \SLVehRightMeanIL{}\,m and RL at \SLVehRightMeanRL{}\,m. \Cref{fig:supp_vehicle_frames} shows the front camera when the safety driver took over. The videos on the project page show both drives of every scored pair with the same overlays. On both curves and in both versions, the planned trajectory leaves the route toward the inner curb.

\rhead{Vision-language judges on drive videos.}
Qwen2.5-VL-3B, Qwen2.5-VL-7B and Qwen3-VL-8B each read 12 front-camera frames of every scored drive, evenly spaced from the start. All clips of a route have one length and end 1\,s before the earliest intervention on that route (\SLVehClipLeft{}\,s on the main route), so no clip shows an intervention. The judges saw no label and gave the probability that the drive would need an intervention. Against the records of the \SLVehScored{} scored drives their AUROC is \SLVehAurocJTwo{}, \SLVehAurocJThree{} and \SLVehAurocJFour{} (\cref{fig:supp_vehicle_judges}): all three models rated the straight drives, which were all completed, as more likely to need an intervention than most curve drives. On the main route no judge separates the versions: the mean probabilities they gave IL and RL differ by at most \SLVehPrefMaxDiff{}.

\section{Survey of self-improvement loops}
\label{sec:supp_survey}

\rhead{Surveyed loops.}
\Cref{tab:prior} shows a selection of the \SLSurveyLoopsSThree{} embodied self-improvement loops we surveyed, \SLSuppSurveyManipN{} in robot manipulation (\cref{tab:supp_survey_manip}) and \SLSuppSurveyDriveN{} in driving (\cref{tab:supp_survey_drive}). The survey takes the term broadly: it covers loops with several rounds, single post-training steps and refinement at inference. For each loop we recorded the judge that decides inside it, the number of rounds it reports, whether it gives a result for every generation, and whether it measures a case of the policy exploiting its judge. Of these loops, \SLSurveyCurveCompleteSThree{} report a complete per-generation curve and \SLSurveyCurvePartialSThree{} a partial one, \SLSurveyJudgeErrorQuantifiedSThree{} quantify the error of their judge, and \SLSurveyBothAndGamedSThree{} report both a curve and a measured case of exploitation. \Cref{tab:supp_survey_other} adds the two loops outside embodied AI that report a loss after the peak.

\renewcommand{\dblfloatpagefraction}{0.9}
\begin{table*}[tp]
\centering\footnotesize
\setlength{\tabcolsep}{4pt}
\begin{tabular*}{\textwidth}{@{\extracolsep{\fill}}lllll@{}}
\toprule
Loop & Judge inside the loop & Rounds reported & Per-gen.\ curve & Judge exploitation \\
\midrule
SimpleVLA-RL~\cite{ax2509_09674} & simulator environment; real-robot physical task completion & n.s. & no & no \\
$\pi_{\mathrm{RL}}$~\cite{ax2510_25889} & simulator & n.s. & no & no \\
ConRFT~\cite{ax2502_05450} & binary classifier + human interventions & n.s. & no & mentioned \\
RLDG~\cite{ax2412_09858} & HIL-SERL classifier & n.s. & no & no \\
$\pi^{*}_{0.6}$ (RECAP)~\cite{recap2025} & human raters & 2 & partial & no \\
RL Token~\cite{ax2604_23073} & human; per episode & n.s. & no & no \\
VLAC~\cite{ax2509_15937} & VLM process reward + done signal & n.s. & no & no \\
Robo-Dopamine~\cite{ax2512_23703} & reward-model threshold & n.s. & no & described \\
HIL-SERL~\cite{ax2410_21845} & classifier (200 positive / 1000 negative) + human & n.s. & no & no \\
SiLRI~\cite{ax2512_24288} & unassisted completion & n.s. & no & no \\
RoboCat~\cite{ax2306_11706} & reward model trained on crowd-sourced labels & n.s. & no & judge error only \\
Q-Planning~\cite{ax2608_21204} & human; per episode & 5 & partial & described \\
SARM2/SPIRAL~\cite{ax2606_10305} & stage reward model; final evaluation by humans & 3 & partial & mentioned \\
Human-Video Dynamics~\cite{ax2606_21406} & human-labeled outcomes & n.s. & no & no \\
ENPIRE~\cite{ax2606_19980} & sensor-fusion synthetic criterion & n.s. & partial & no \\
DreamGen~\cite{ax2505_12705} & 10 real-robot trials per checkpoint & n.s. & no & judge error only \\
WMPO~\cite{ax2511_09515} & 30 real-robot trials & n.s. & partial & judge error only \\
VLAW~\cite{ax2602_12063} & fine-tuned Qwen3-VL; threshold 0.8 & 2 & partial & judge error only \\
RISE~\cite{ax2602_11075} & milestone scoring; 20 trials & n.s. & no & no \\
WorldSample~\cite{ax2607_02431} & goal condition & n.s. & no & no \\
E2HiL~\cite{ax2601_19969} & human + classifier & n.s. & no & no \\
GenSim2~\cite{ax2410_03645} & solver success rate & n.s. & no & no \\
FATE~\cite{ax2603_01505} & RoboBrain 2.0 + simulated execution & n.s. & no & judge error only \\
RoboReward~\cite{ax2601_00675} & human vs VLM comparison & n.s. & no & judge error only \\
\bottomrule
\end{tabular*}

\caption{\textbf{Surveyed loops in robot manipulation.} Judge exploitation: whether the paper mentions, describes or measures the policy exploiting its judge; ``judge error only'' marks papers that quantify the judge's error without such a case. n.s.: the paper states no number of rounds.}
\label{tab:supp_survey_manip}
\end{table*}

\begin{figure*}[tp]
\centering
\includegraphics[width=\textwidth]{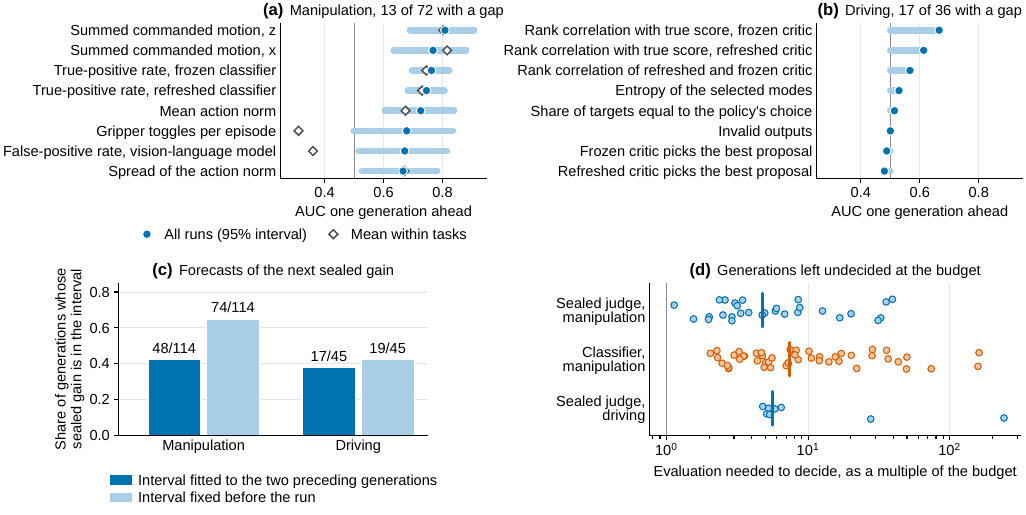}
\caption{\textbf{Warning signals, forecasts and undecided generations.} (a, b) AUC with which a signal of the preceding generation predicts that the judge gap of a generation reaches the threshold; the eight best signals of each testbed. (c) Share of generations whose sealed gain lies in the interval forecast for it. (d) Evaluation that the test would need to decide each generation it left undecided, as a multiple of the budget it had; bars mark medians.}
\label{fig:supp_additional}
\end{figure*}

\begin{table*}[t]
\centering\footnotesize
\setlength{\tabcolsep}{4pt}
\begin{tabular*}{\textwidth}{@{\extracolsep{\fill}}lllll@{}}
\toprule
Loop & Judge inside the loop & Rounds reported & Per-gen.\ curve & Judge exploitation \\
\midrule
GIGAFLOW~\cite{ax2502_03349} & simulator rules + WOSAC & 1; about $10^{12}$ transitions & no & judge error only \\
Cornelisse self-play~\cite{ax2502_14706} & simulator rules & 1; $2{\times}10^{9}$ steps & no & no \\
SPACeR~\cite{ax2510_18060} & reference-model likelihood + rules & 1 & no & described \\
Scaling Self-Play~\cite{ax2606_19641} & privileged teacher + rules & DAgger; $1.5{\times}10^{8}$ steps & no & no \\
TerraTransfer~\cite{ax2606_17386} & self-play teacher & 1 & no & no \\
RAD~\cite{ax2502_13144} & 3DGS rule-based reward & 1 & no & no \\
RAD-2~\cite{ax2604_15308} & learned discriminator & 1 & no & no \\
CaRL~\cite{ax2504_17838} & route-completion reward & 1; $3{\times}10^{8}$ / $5{\times}10^{8}$ samples & no & described \\
PerlAD~\cite{ax2603_14908} & pseudo-simulation rules + world model & 1 & no & judge error only \\
MindDrive~\cite{ax2512_13636} & sparse rule-based reward & 2 online epochs & no & no \\
ReCogDrive~\cite{ax2506_08052} & NAVSIM simulator PDMS & 1 & no & no \\
DiffusionDriveV2~\cite{ax2512_07745} & trajectory-level rule-based reward & 1 & no & no \\
DriveDPO~\cite{ax2509_17940} & rule-based PDMS preferences & 1 DPO stage & no & no \\
SMART-R1~\cite{ax2509_23993} & WOSAC meta-metric & one SFT-RFT-SFT cycle; per stage & partial & no \\
ActiveAD~\cite{ax2403_02877} & displacement / soft collision / uncertainty & 2; per-round readings & \good yes & no \\
Mcity Data Engine~\cite{ax2504_21614} & open-vocabulary ensemble & 1 & no & judge error only \\
R2SE~\cite{ax2506_09800} & PDMS + loss + entropy & 1 & no & no \\
AutoScale~\cite{ax2605_21372} & closed-loop evaluation feedback & $T{=}5$; final value only & no & no \\
R$^2$LPL~\cite{ax2606_30537} & rule-based detectors & 5 (plus a 10-round envelope) & \good yes & no \\
ELF-VLA~\cite{ax2603_01063} & Qwen3-VL-32B diagnosis & $k{=}1$ & no & mentioned \\
SceneCrafter~\cite{ax2503_18108} & 3DGS simulator & 1 & no & no \\
SimScale~\cite{ax2511_23369} & pseudo-expert + rules & 1 & no & no \\
Cosmos-Drive-Dreams~\cite{ax2506_09042} & none (supervised) & 1 & no & no \\
World Engine~\cite{ax2606_19836} & rules in reconstructed environments & 1; multiple rounds unstable & no & no \\
AD-R1~\cite{ax2511_20325} & impartial world-model critic & 1 & no & judge error only \\
ReSim~\cite{ax2506_09981} & Video2Reward & 1 & no & judge error only \\
EvaDrive~\cite{ax2508_09158} & learned multi-objective critic & $\geq$4; best at 2 to 3 & partial & no \\
CriticVLA~\cite{ax2604_27366} & critic trained on rule-based labels & multi-step refinement; diminishing & partial & no \\
DriveReward~\cite{ax2606_08525} & 1B VLM reward model & 1 & no & no \\
Learning to Drive in a Day~\cite{ax1807_00412} & safety-driver takeover & 11 to 35 episodes & no & no \\
H-DSAC~\cite{ax2510_06038} & human interventions & $10^{5}$ steps & no & no \\
Waymo (blog post, Dec.\ 2025) & simulator + critic & n.s. & no & no \\
\bottomrule
\end{tabular*}

\caption{\textbf{Surveyed loops in driving.} Columns as in \cref{tab:supp_survey_manip}.}
\label{tab:supp_survey_drive}
\end{table*}

\section{Additional analyses}
\label{sec:supp_additional}

\begin{table}[t]
\centering\footnotesize
\setlength{\tabcolsep}{3pt}
\begin{tabular}{@{}>{\raggedright\arraybackslash}p{0.31\linewidth}>{\raggedright\arraybackslash}p{0.23\linewidth}>{\raggedright\arraybackslash}p{0.36\linewidth}@{}}
\toprule
Loop & Judge & Reported effect \\
\midrule
Self-Regress~\cite{ax2606_21090} (competitive programming) & CodeGrader & collapse of 17 points after the peak \\
Verifiers Go Bad~\cite{ax2606_14629} (multimodal QA) & frozen VLM verifier & $-$3.4 to $-$10.9 points across tasks \\
\bottomrule
\end{tabular}

\caption{\textbf{Loops outside embodied AI.} Both report a loss from further rounds of self-improvement.}
\label{tab:supp_survey_other}
\end{table}

\begin{table}[t]
\centering\footnotesize
\setlength{\tabcolsep}{1.7pt}
\setbox0\hbox{\begin{tabular}{@{}l c r cc r@{}}
\toprule
 & Learning & & \multicolumn{2}{c}{Success} & \\
\cmidrule(lr){4-5}
Training data & rate & $n$ & Start & Continued & Change \\
\midrule
Own successes & $5{\times}10^{-4}$ & 60 & 0.633 & 0.383 & ${-}0.25$ [${-}0.44$, ${-}0.06$] \\
Own successes & $1{\times}10^{-4}$ & 60 & 0.633 & 0.500 & ${-}0.13$ [${-}0.30$, $+0.03$] \\
Own successes & $3{\times}10^{-5}$ & 60 & 0.633 & 0.333 & ${-}0.30$ [${-}0.44$, ${-}0.16$] \\
\midrule
Own successes & $1{\times}10^{-4}$ & 120 & 0.675 & 0.408 & ${-}0.27$ [${-}0.38$, ${-}0.16$] \\
Demonstrations & $1{\times}10^{-4}$ & 120 & 0.675 & 0.983 & $+0.31$ [$+0.23$, $+0.39$] \\
Both, 1:1 & $1{\times}10^{-4}$ & 120 & 0.675 & 0.733 & $+0.06$ [${-}0.06$, $+0.17$] \\
\bottomrule
\end{tabular}
}\addtolength{\tabcolsep}{\dimexpr(\linewidth-\wd0)/10\relax}
\begin{tabular}{@{}l c r cc r@{}}
\toprule
 & Learning & & \multicolumn{2}{c}{Success} & \\
\cmidrule(lr){4-5}
Training data & rate & $n$ & Start & Continued & Change \\
\midrule
Own successes & $5{\times}10^{-4}$ & 60 & 0.633 & 0.383 & ${-}0.25$ [${-}0.44$, ${-}0.06$] \\
Own successes & $1{\times}10^{-4}$ & 60 & 0.633 & 0.500 & ${-}0.13$ [${-}0.30$, $+0.03$] \\
Own successes & $3{\times}10^{-5}$ & 60 & 0.633 & 0.333 & ${-}0.30$ [${-}0.44$, ${-}0.16$] \\
\midrule
Own successes & $1{\times}10^{-4}$ & 120 & 0.675 & 0.408 & ${-}0.27$ [${-}0.38$, ${-}0.16$] \\
Demonstrations & $1{\times}10^{-4}$ & 120 & 0.675 & 0.983 & $+0.31$ [$+0.23$, $+0.39$] \\
Both, 1:1 & $1{\times}10^{-4}$ & 120 & 0.675 & 0.733 & $+0.06$ [${-}0.06$, $+0.17$] \\
\bottomrule
\end{tabular}

\caption{\textbf{Training data of the improver.} A start policy of task 1 trained for \SLAddCalibSteps{} more steps on each kind of data. Demonstrations: those used to adapt the start policy. Change: paired change in success rate with 95\% interval.}
\label{tab:supp_calibration}
\end{table}

\subsection{Anticipating the judge gap}
\rhead{Early signs of the judge gap.}
In \SLGapPositiveChainsSOne{} classifier-decided manipulation runs, the 95\% interval of the judge gap lies above zero in at least one of the first five generations. We asked whether quantities that a loop can log without a sealed judge predict this one generation ahead. For generations two to five of every run, the signals of the preceding generation are paired with the event that the judge gap reaches \SLGapThresholdSOne{} on manipulation (\SLAddWarnManipPos{} of \SLAddWarnManipPairs{} pairs) or \SLGapThresholdSThree{} on driving (\SLAddWarnDrivePos{} of \SLAddWarnDrivePairs{}). The signals are statistics of the policy's actions, error rates of the judges and, on driving, statistics of the critic's picks. \Cref{fig:supp_additional}a,b give the AUC of each. On manipulation \SLAddWarnManipAbove{} of \SLAddWarnManipFeatures{} signals reach 0.6. The best is the summed commanded motion along the vertical axis, with \SLAddWarnManipTop{} \SLAddWarnManipTopCI{} and \SLAddWarnManipTopWithin{} within tasks. Not every signal survives this control: the number of gripper toggles falls from \SLAddWarnManipToggles{} to \SLAddWarnManipTogglesWithin{} within tasks, so it separates tasks, not generations. On driving only \SLAddWarnDriveAbove{} of \SLAddWarnDriveFeatures{} signals reach 0.6, and the best, the frozen critic's rank correlation with the true score, reaches \SLAddWarnDriveTop{}. Changes in the policy's motion thus carry some warning on manipulation, and none of the \SLAddWarnDriveFeatures{} signals we logged does on driving.

\begin{table*}[t]
\centering\footnotesize
\setlength{\tabcolsep}{5pt}
\setbox0\hbox{\begin{tabular}{@{}l c c lr lr r r cc@{}}
\toprule
Deciding & & & \multicolumn{2}{c}{Refreshed} & \multicolumn{2}{c}{Frozen} & Sealed & & \multicolumn{2}{c}{FPR} \\
\cmidrule(lr){4-5}\cmidrule(lr){6-7}\cmidrule(l){10-11}
judge & Task & Gen. & Decision & Pairs & Decision & Pairs & gain & Failures & Refreshed & Frozen \\
\midrule
Classifier & 2 & 4 & Undecided & 300 & Reject & 161 & ${-}0.153$ & 89 & 0.46 & 0.10 \\
 & 2 & 5 & Undecided & 300 & Reject & 13 & ${-}0.350$ & 97 & 0.53 & 0.07 \\
 & 3 & 4 & Undecided & 300 & Reject & 54 & ${-}0.136$ & 36 & 1.00 & 0.00 \\
 & 3 & 6 & Undecided & 300 & Reject & 35 & ${-}0.226$ & 68 & 1.00 & 0.00 \\
\midrule
Sealed judge & 2 & 4 & Accept & 188 & Undecided & 300 & $+0.109$ & 19 & 0.47 & 0.21 \\
 & 3 & 4 & Undecided & 300 & Reject & 270 & ${-}0.035$ & 42 & 0.00 & 0.00 \\
 & 3 & 5 & Undecided & 300 & Accept & 162 & $+0.082$ & 35 & 0.00 & 0.03 \\
\bottomrule
\end{tabular}
}\addtolength{\tabcolsep}{\dimexpr(\textwidth-\wd0)/20\relax}
\begin{tabular}{@{}l c c lr lr r r cc@{}}
\toprule
Deciding & & & \multicolumn{2}{c}{Refreshed} & \multicolumn{2}{c}{Frozen} & Sealed & & \multicolumn{2}{c}{FPR} \\
\cmidrule(lr){4-5}\cmidrule(lr){6-7}\cmidrule(l){10-11}
judge & Task & Gen. & Decision & Pairs & Decision & Pairs & gain & Failures & Refreshed & Frozen \\
\midrule
Classifier & 2 & 4 & Undecided & 300 & Reject & 161 & ${-}0.153$ & 89 & 0.46 & 0.10 \\
 & 2 & 5 & Undecided & 300 & Reject & 13 & ${-}0.350$ & 97 & 0.53 & 0.07 \\
 & 3 & 4 & Undecided & 300 & Reject & 54 & ${-}0.136$ & 36 & 1.00 & 0.00 \\
 & 3 & 6 & Undecided & 300 & Reject & 35 & ${-}0.226$ & 68 & 1.00 & 0.00 \\
\midrule
Sealed judge & 2 & 4 & Accept & 188 & Undecided & 300 & $+0.109$ & 19 & 0.47 & 0.21 \\
 & 3 & 4 & Undecided & 300 & Reject & 270 & ${-}0.035$ & 42 & 0.00 & 0.00 \\
 & 3 & 5 & Undecided & 300 & Accept & 162 & $+0.082$ & 35 & 0.00 & 0.03 \\
\bottomrule
\end{tabular}

\caption{\textbf{Generations that the refreshed and the frozen classifier decide differently.} The classifier's test replayed on the same pairs with the verdicts of the classifier refreshed in the loop and of the classifier frozen before it. Failures: failed episodes of the candidate among the \SLEvalPerGenInLoopSOne{} pairs; FPR: share of them each classifier labels successful.}
\label{tab:supp_flips}
\end{table*}

\rhead{Forecasting the sealed gain.}
Before each sealed reading we fitted an interval for the sealed gain to the two preceding generations of the run. The measured gain fell in it in \SLFTwoHitRateSOne{} manipulation and \SLFTwoHitRateSThree{} driving generations (\cref{fig:supp_additional}c). An interval fixed before the run and never updated contains \SLFTwoPriorBandThroughoutSOne{} and \SLFTwoPriorBandThroughoutSThree{}. In these runs the interval fitted to the two preceding generations does no better than the fixed one.

\subsection{The sealed judge and its budget}
\rhead{A vision-language model as the sealed judge.}
The sealed judge of the manipulation runs is the simulator's success predicate, which a robot outside simulation does not have. We tested Qwen2.5-VL-7B as its stand-in, with one threshold fitted on \SLAddDetFit{} episodes. On \SLAddDetN{} further episodes it reaches a true-positive rate of \SLAddDetTPR{} at a false-positive rate of \SLAddDetFPR{} against the predicate (AUC \SLAddDetAUC{}), so sensitivity plus specificity minus one is \SLAddDetJ{}. The value differs widely between tasks (\SLAddDetJTaskOne{}, \SLAddDetJTaskTwo{} and \SLAddDetJTaskThree{}). A sealed judge this noisy would add its own error to every acceptance. The perturbed pool of \cref{fig:supp_perturbed} is a second sealed judge that keeps the predicate. Of the four pairs of policies of one task compared on both pools, the test decides one on the perturbed pool, and it orders that pair as the sealed pool does (Kendall's $\tau$ \SLKendallTauSOne{}; \SLKendallTauAllPairsSOne{} over the point estimates of all four). The perturbed pool thus agrees with the sealed pool on the one pair it can decide.

\rhead{Evaluation needed by undecided generations.}
A generation that is undecided at its budget is rolled back. For each one we projected the number of pairs at which the test would decide if the observed trend of the paired differences continued (\cref{fig:supp_additional}d). On manipulation this covers \SLAddNeedManipReadings{} undecided tests, which would need \SLAddNeedManipMin{} to \SLAddNeedManipMax{} further pairs. The median is \SLAddNeedSealedManipMedian{} times the budget for the \SLAddNeedSealedManipN{} tests fed by the sealed judge and \SLAddNeedClassManipMedian{} times for the \SLAddNeedClassManipN{} fed by the classifier. On driving, \SLAddNeedSealedDriveN{} generations of the sealed-decided main and control runs are undecided per log with the whole pool read; they would need a median of \SLAddNeedSealedDriveMedian{} times the logs of the pool and up to \SLAddNeedSealedDriveMax{} times. The undecided candidates are those whose gain is close to zero, and a moderately larger budget would not have decided them.

\subsection{Components of the manipulation loop}
\rhead{Training data of the improver.}
Before the first generation we continued one start policy on three kinds of data and measured it on calibration conditions (\cref{tab:supp_calibration}). Trained on its own successful episodes, the policy loses at each of three learning rates (\SLAddCalibOwnMin{} to \SLAddCalibOwnMax{} on 60 conditions, \SLAddCalibOwnLarge{} on 120). Trained on the original demonstrations it gains \SLAddCalibDemo{}, and on an equal mixture of both \SLAddCalibMix{} \SLAddCalibMixCI{}. Demonstrations replayed through the same path as the collected episodes give \SLExtraDemoPath{} \SLExtraDemoPathCI{} on \SLExtraDemoPathN{} conditions, so the loss comes from the collected episodes and not from the path they take. A loop that replays demonstrations would therefore improve through the demonstrations under either judge, and its gains could not be attributed to the collected data or to the judge that selects them. The improver of all runs trains on the collected episodes alone, at the learning rate with the smallest loss. The loops agree with this: the sealed test kept \SLSealedAcceptedGensSOne{} of the \SLPostBudgetSealedN{} candidates trained this way (mean certified gain \SLSealedGainLastSOne{}) and rolled back the rest.

\rhead{Effect of refreshing the classifier.}
We replayed the classifier's test in all \SLAddFlipsOf{} generations with the verdicts of the classifier frozen before the loop. The decision differs in \SLAddFlips{} (\cref{tab:supp_flips}). In \SLAddFlipsLenient{} of them the refreshed classifier labels more of the candidate's failures as successes than the frozen one, in two generations all of them. In the \SLAddFlipsClassifierRuns{} generations that the classifier decided, the frozen classifier rejects a candidate that the refreshed one leaves undecided, so the run rolls back either way and no kept policy changes. On the 300 collected episodes of a generation the two classifiers disagree on at most one episode in \SLClassifierDriftZeroOrOneSOne{} generations and on no more than \SLClassifierDriftMaxSOne{} in any. On the 300 evaluation episodes of the candidate they disagree on at most one in \SLAddEvalDisagreeZeroOrOne{} generations and on up to \SLAddEvalDisagreeMax{}. The refresh thus changes little on the episodes the classifier is trained on and, in some generations, much on the failures of the next candidate.

\rhead{Dependence on the start state.}
A condition is a start state with an environment seed. The seed changes the trajectory but rarely the outcome: \SLDistinctTrajectoriesSOne{} recorded episodes are distinct action sequences, yet in \SLSeedOutcomeAgreementSOne{} groups of one run and one start state all six seeds end alike. In the sealed records, \SLPairedOutcomeIdenticalSOne{} groups of one generation and one start state have the same paired outcome on every seed drawn. The pairs of a test are therefore correlated within a start state; \cref{sec:supp_method} measures the error rate under this sampling, and \cref{fig:supp_audit} tests a certified gain on new start states.

\section{Limitations}
\label{sec:supp_limitations}

Our loops run in simulation, on three short-horizon manipulation tasks and one driving benchmark, each with one published improver. The driving benchmark replays recorded logs, so other road users do not react to the policy. The field experiment tests one published improvement step on one vehicle, on routes without obstacles. Longer tasks, on-policy reinforcement learning and loops that run on hardware over several generations are left to future work.

\end{document}